\pdfoutput=1

\documentclass[11pt]{article}

\usepackage[preprint]{acl}

\usepackage{times}
\usepackage{latexsym}

\usepackage[T1]{fontenc}

\usepackage[utf8]{inputenc}

\usepackage{microtype}

\usepackage{inconsolata}

\usepackage{graphicx}
\usepackage{multirow}
\usepackage{booktabs}
\usepackage{amsmath}
\usepackage[most]{tcolorbox} 
\usepackage{enumitem}
\usepackage{xcolor}

\tcbset{
    mybox/.style={
        colframe=darkgray,          
        colback=darkgray!10,        
        coltitle=white,         
        colbacktitle=darkgray,      
        boxrule=0.5mm,          
        arc=3mm,                
        width=\linewidth,       
        left=1mm,               
        right=1mm,              
        top=1mm,                
        bottom=1mm,             
    }
}

\title{How Do LLMs Acquire New Knowledge?\\ A Knowledge Circuits Perspective on Continual Pre-Training
}

\author{
Yixin Ou{$^{\spadesuit}$},
Yunzhi Yao{$^{\spadesuit}$},
Ningyu Zhang{$^{\spadesuit} \thanks{~Corresponding Author.}$},
\textbf{Hui Jin}{$^{\heartsuit}$},\\
{\bf Jiacheng Sun}{$^{\heartsuit}$},
{\bf Shumin Deng}{$^{\clubsuit}$},
{\bf Zhenguo Li}{$^{\heartsuit}$},
{\bf Huajun Chen}$^{\spadesuit\diamondsuit}$\\
 $^\spadesuit$Zhejiang University
 $^\heartsuit$Huawei Noah’s Ark Lab \\
 $^\clubsuit$National University of Singapore, NUS-NCS Joint Lab, Singapore \\
 $^\diamondsuit$Zhejiang Key Laboratory of Big Data Intelligent Computing\\
  \texttt{\{ouyixin,zhangningyu,huajunsir\}@zju.edu.cn}\\
}

\begin{document}
\maketitle

\begin{abstract}
Despite exceptional capabilities in knowledge-intensive tasks, Large Language Models (LLMs) face a critical gap in understanding how they internalize new knowledge, particularly how to structurally embed acquired knowledge in their neural computations. We address this issue through the lens of knowledge circuit evolution, identifying computational subgraphs that facilitate knowledge storage and processing. Our systematic analysis of circuit evolution throughout continual pre-training reveals several key findings: (1) the acquisition of new knowledge is influenced by its relevance to pre-existing knowledge; (2) the evolution of knowledge circuits exhibits a distinct phase shift from formation to optimization; (3) the evolution of knowledge circuits follows a deep-to-shallow pattern. These insights not only advance our theoretical understanding of the mechanisms of new knowledge acquisition in LLMs, but also provide potential implications for improving continual pre-training strategies to enhance model performance\footnote{Code and data are available at \url{https://github.com/zjunlp/DynamicKnowledgeCircuits}.}.
\end{abstract}

\section{Introduction}

Knowledge is a cornerstone of intelligence, shaping how humanity perceives the world, interacts with others, and navigates daily life \citep{DBLP:conf/wsdm/Choi22,lkm}.
As human society advances, the ways by which knowledge is stored, accessed, and processed have evolved significantly, especailly with the advent of Large Language Models (LLMs).
Recent studies \citep{gpt3,gpt4,llama3,deepseek_v3,qwen2.5,llm_survey,DBLP:journals/corr/abs-2401-10034} on LLMs have demonstrated their ability to capture factual knowledge from pre-training corpus and encapsulate it as extensive parametric knowledge, empowering their remarkable capabilities in numerous knowledge-intensive tasks \citep{knowledge_mechanisms_survey,knowledge_life_cycle}, as well as in developing higher-order capabilities like reasoning \citep{lm_prompt_reasoning_survey,llm_reasoning_survey}.
Nevertheless, these powerful models still struggle with knowledge updates, especially with regard to the dynamic nature of world knowledge that evolves after the cut-off date of the pre-training corpus \citep{DBLP:conf/emnlp/ZhangFCNW23,DBLP:journals/corr/abs-2404-08700}.
Extensive efforts focus on developing advanced techniques for injecting new knowledge into LLMs \citep{continual_knowledge_learning,pit,sft_injecting,DBLP:conf/emnlp/OvadiaBME24,DBLP:journals/corr/abs-2411-07175}, yet the absence of a well-defined mechanism for new knowledge acquisition in LLMs continues to hinder further progress in this area.

\begin{figure}
    \centering
    \includegraphics[width=\linewidth]{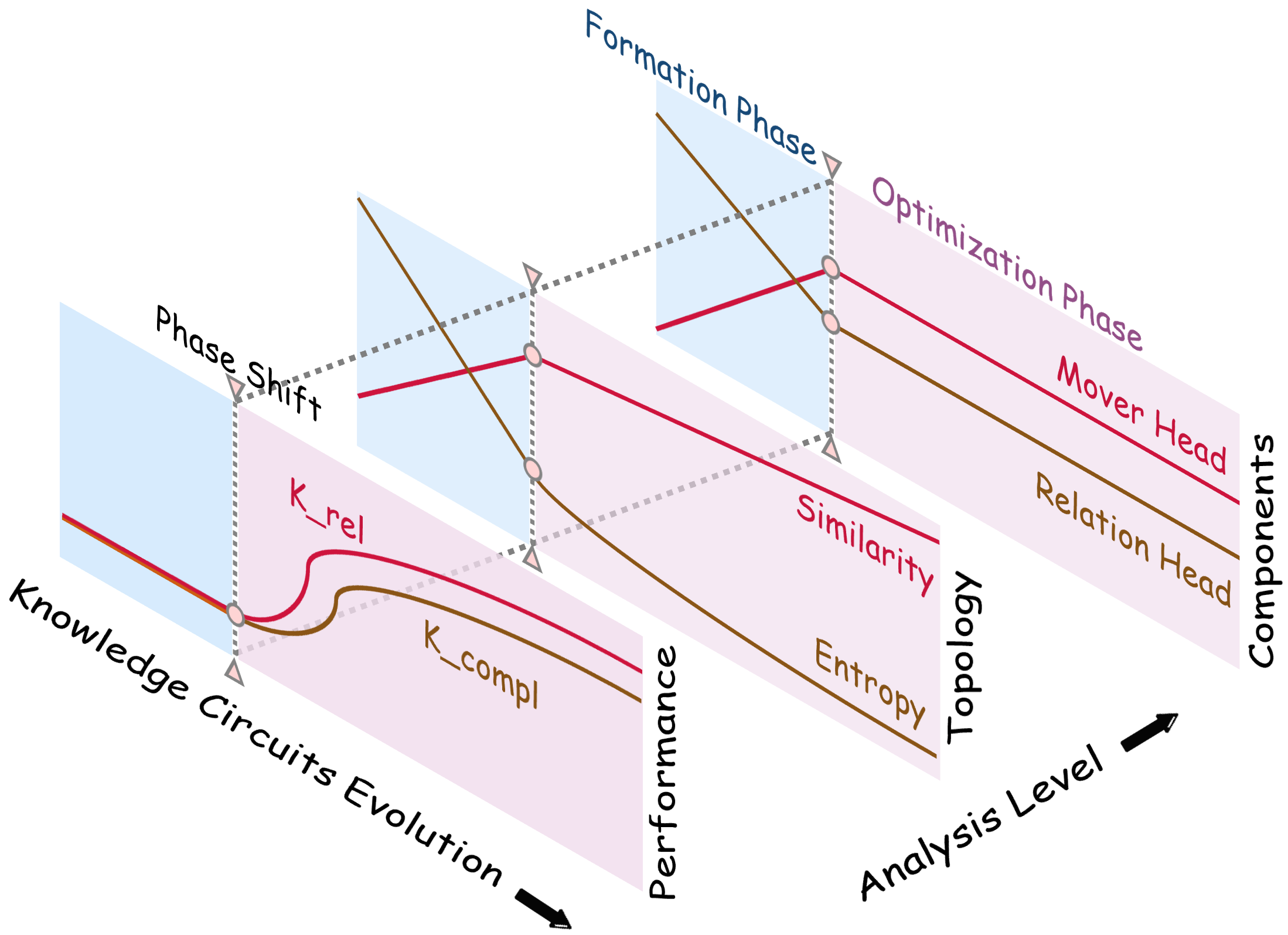}
    \caption{Illustration of our findings: \textbf{Phase shift} from formation to optimization in the evolution of knowledge circuits, each phase characterized by distinct features at the performance, topology, and component levels.}
    \label{fig:illustration}
\end{figure}

Recent works introduce mechanistic interpretability techniques to uncover knowledge machanisms in LLMs.
\citet{physics3.1} adopts probing methods to examine the storage and extraction of factual knowledge encoded in hidden states of language models.
\citet{knowledge_entropy} introduces the concept of knowledge entropy to examine how the integration of knowledge of LLMs evolves during the pre-training phase.
However, previous works typically treat knowledge blocks as isolated components and often focus on identifying specific blocks that store particular knowledge.
In contrast, \citet{knowledge_circuits} move beyond isolated components and explore the computation graph to uncover knowledge circuits, investigating cooperation between different components to understand how knowledge is stored and expressed.

In this paper, we investigate the mechanism of new knowledge acquisition in LLMs from the perspective of knowledge circuits.
By analyzing the evolution of knowledge circuits throughout continual pre-training, we uncover several interesting findings, as illustrated in Figure \ref{fig:illustration}.

Key findings of the paper are summarized as:
\begin{itemize}[itemsep=0pt, topsep=5pt]
    \item (\S\ref{sec:performance}) The acquisition of new knowledge is significantly influenced by its relevance to pre-existing knowledge, with relevant new knowledge being integrated more efficiently than completely new knowledge.
    \item (\S\ref{sec:topology}) In the process of knowledge acquisition, the evolution of knowledge circuits exhibits a distinct phase shift from formation to optimization, each marked by unique structural and behaviral characteristics.
    \item (\S\ref{sec:components}) The evolution of knowledge circuits follows a deep-to-shallow pattern, where mid-to-deeper layers first develop the extraction function, and later, lower layers enrich their knowledge representations.
\end{itemize}
These findings offer valuable insights into the mechanisms by which LLMs adapt their internal structures to acquire new knowledge.
This understanding not only informs potential strategies for enhancing the continual learning capabilities of LLMs but also provides a solid foundation for improving their adaptability across diverse domains.

\section{Background}

\subsection{Circuit Theory}

\paragraph{Circuit as Computational Subgraph}
Delving into the Transformer architecture~\citep{transformer}, all computations in a Transformers-based language model as a connected directed acyclic graph, denoted as $\mathcal{G}$.
This graph represents the flow of information from the input of the language model to the token unembedding, where activations are projected back to vocabulary space.
Various components of a language model, including attention heads and multi-layer perceptrons (MLPs), are defined as the nodes of this graph, denoted as $N$.
The edges of this graph, denoted as $E$, are the weighted connections between these components, encompassing residual connections, attention mechanisms, and projections.
In the context of Mechanistic Interpretability (MI), which aims to understand the inner workings of advanced Transformer-based language models~\citep{mi_review,mi_primer,mi_safety_review,sharkey2025open}, a \textbf{circuit} is conceptualized as a sparse computational subgraph $\mathcal{C}\subset\mathcal{G}$ within a language model whose computations are most relevant to the whole model's behaviour on the specific task~\citep{zoom_in,transformer_circuits,interpretability_in_the_wild,sparse_feature_circuits}.
A circuit $\mathcal{C}$ usually contains a selection of nodes $N_{\mathcal{C}}\subset{N}$ and edges $E_{\mathcal{C}}\subset{E}$ necessary for the specific task, expressed as $\mathcal{C}=<N_{\mathcal{C}},E_{\mathcal{C}}>$.

\paragraph{Circuit Discovery}
The goal of circuit discovery is to identify a computational subgraph that represents the whole model's behavior on a specific task.
Many studies adopt causal mediation analysis to localize critical nodes or edges within language models in order to identify and verify circuits.
\citet{acdc} adopts activation patching and proposes ACDC.
\citet{eap} introduces Edge Attribution Patching (EAP) to make a linear approximation of
activation patching, which assigns an importance score to each edge.

\subsection{Knowledge Circuits}
Unlike previous works~\citep{knowledge_neurons,ffn_kv,llm_factual_recall,rome} that treat the knowledge blocks as isolated components, ~\citet{knowledge_circuits} introduce a novel perspective: knowledge circuits.
They hypothesize that the cooperation between multiple components unveils the implicit knowledge representation in LLMs.
An identified knowledge circuit is considered a computational subgraph that faithfully represents specific knowledge domains within the model's parametric memory.
As such, it should be capable of independently reproducing the behavioral patterns or performance of the entire model with respect to the corresponding tasks.
However, ~\citet{knowledge_circuits} concentrates exclusively on the knowledge that already stored in the language model, without investigating the process by which LLMs acquire knowledge.
In this work, we aim to advance the concept of knowledge circuits by investigating their dynamics throughout continual pre-training. 

\section{Methodology}

\subsection{Dataset Construction}
\label{sec:dataset}

Given the challenges of conducting mechanistic interpretability analysis on Internet-scale corpus, we perform controlled experiments on synthetic data, following~\citet{physics3.1,physics3.2,physics3.3}.
We focus on factual knowledge that can be represented as triples of the form $(s, r ,a)$ containing subject $s$, relation $r$, and attribute $a$.
We synthesize a pool of fictional knowledge entities based on heuristic rules using ChatGPT, ensuring that these fictional biographical knowledge is unavailable to LLMs in the pre-training phase.
Each knowledge entity is first assigned a unique name as the subject, and then associated with five relations—\textit{birth date}, \textit{city}, \textit{major}, \textit{university} and \textit{company}-and corresponding attributes.
To convert these entities into textual knowledge for training data, we fill them in predefined templates.
Considering real-world data scenarios and the perspectives of analysis, we further customize the training corpus from two aspects: \textit{knowledge type} and \textit{knowledge frequency}.

\paragraph{Knowledge Type}
We classify the new knowledge that the language model may need to acquire into two categories. 
One involves knowledge that already exists in the model's parameters but requires further learning of specific aspects (e.g., new relations).
This type of knowledge is referred to as \textit{relevant new knowledge} and denoted as $K_\text{rel}$.
The other type of knowledge is absent from the model's parameters, which is referred to as \textit{completely new knowledge} and denoted as $K_\text{compl}$.

\paragraph{Knowledge Frequency}
Considering the long-tail distribution of knowledge in real-world data, we model the frequency of knowledge entities in the corpus to follow an exponential distribution.
This ensures that the corpus for continual pre-training contains both high-frequency knowledge as well as long-tail knowledge.

More details of the pipeline of dataset construction are provided in Appendix \ref{app:dataset}.

\subsection{Model Training}

To conduct the knowledge acquisition experiment, we use three series of typical decoder-only LLMs to yield consistent findings on different architectures: \textit{GPT-2}, \textit{Llama}, and \textit{Phi}.
We continually pre-train the base models using a standard next-token prediction objective on the corpus described in Section~\ref{sec:dataset}.
Further details on the training configuration can be found in Appendix \ref{app:training}.

\subsection{Circuit Discovery}
\label{sec:circuit_discovery}

To facilitate the discovery of circuits over multiple checkpoints throughout continual pre-training, we select EAP-IG~\citep{eap-ig} from a range of circuit discovery techniques ~\citep{acdc,eap,information_flow_routes,eap-ig}, which assigns an importance score to each edge, balancing efficiency and faithfulness.
Given an edge $e=(u,v)\in{E}$ between nodes $u\in{N}$ and $v\in{N}$ with clean and corrupted activations $z_u$ and $z'_u$, EAP-IG scores the importance of $e$ as:
\begin{equation}
    S(e)=\left(z_u^{\prime}-z_u\right) \frac{1}{m} \sum_{k=1}^m \frac{\partial L\left(z^{\prime}+\frac{k}{m}\left(z-z^{\prime}\right)\right)}{\partial z_v}
    \label{eq:eap-ig}
\end{equation}
where $z$ refers to a sequence of token embeddings for one input, $z'$ refers to the token embeddings of the distinct, baseline input, and $m$ refers to the number of integrated gradient steps; we set $m=5$ as suggested by \citet{eap-ig}.
More details of circuit discovery are provided in Appendix \ref{app:circuit_discovery}.

After scoring all edges within a language model using EAP-IG, we identify a circuit by selecting the top $n$ edges with the highest absolute score as in \citet{eap}, ensuring that the selected edges collectively achieve over 70\% of the whole model's performance on the specific task.
Specifically, we retain 8k, 20k, 50k, and 50k edges for GPT-2 Small, GPT-2 Medium, TinyLlama, and Phi-1.5, respectively.


\section{Analyzing the Evolution of Knowledge Circuits throughout Training}

\begin{figure*}[htp]
    \centering
    \includegraphics[width=\linewidth]{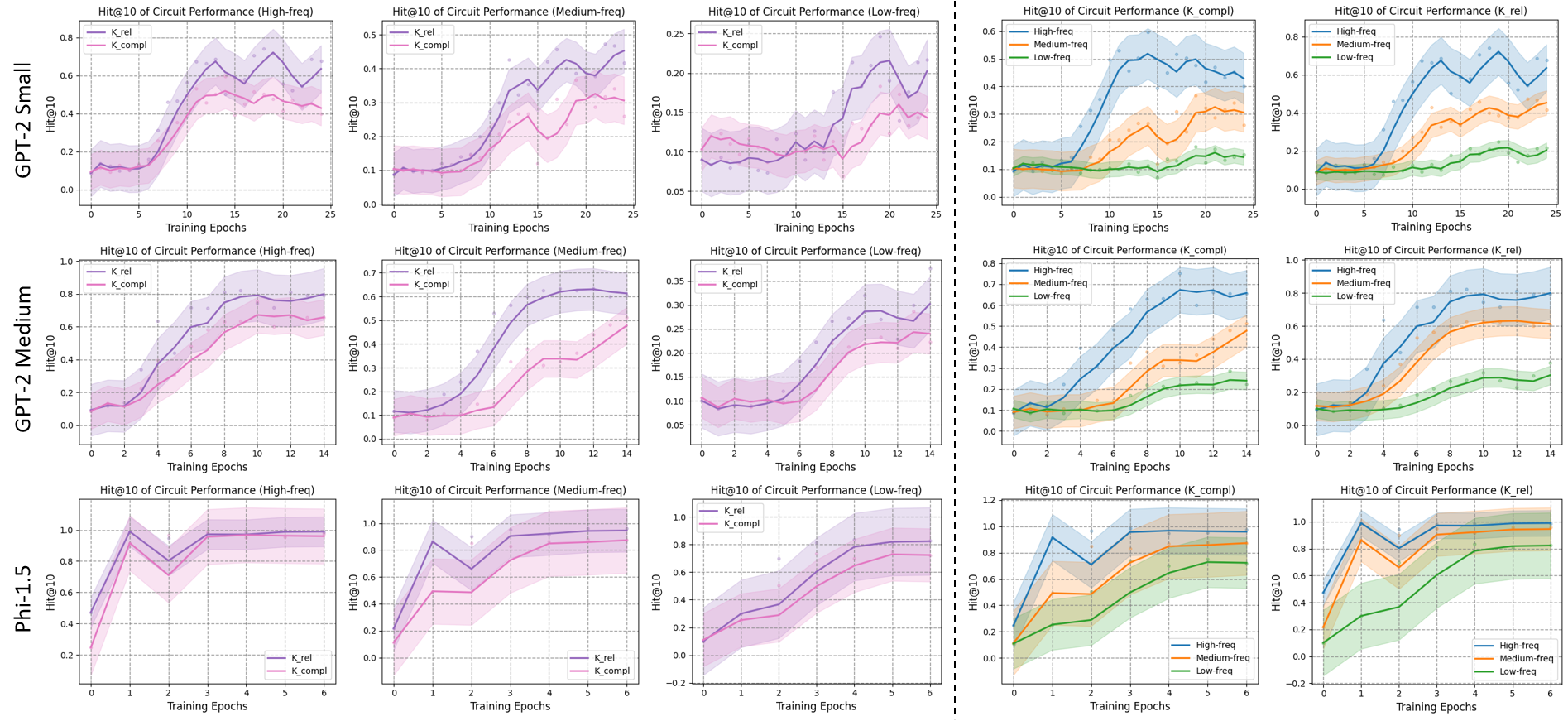}
    \caption{\textbf{Hit@10} of the performance of knowledge circuits in GPT-2 Small, GPT-2 Medium and Phi-1.5 throughout training. Left: Performance for circuits discovered by different types of knowledge, where \textcolor[RGB]{148,103,189}{\texttt{K\_rel}} and \textcolor[RGB]{227,119,194}{\texttt{K\_compl}} represent \textcolor[RGB]{148,103,189}{relevant new knowledge} and \textcolor[RGB]{227,119,194}{completely new knowledge}, respectively. Right: Performance for circuits discovered by different frequencies of knowledge, where \textcolor[RGB]{44,160,44}{\texttt{Low-freq}}, \textcolor[RGB]{255,127,14}{\texttt{Medium-freq}}, and \textcolor[RGB]{31,119,180}{\texttt{High-freq}} represent knowledge with frequencies in the ranges  \textcolor[RGB]{44,160,44}{$[1, 2)$}, \textcolor[RGB]{255,127,14}{$[2,5]$} and \textcolor[RGB]{31,119,180}{$(5, 27]$}, respectively. Note that we smooth the curves using a window size of 3 epochs for all settings.}
    \label{fig:hit_at_10}
\end{figure*}

Once we have identified the knowledge circuits, we delve deeper into the changes within the circuits, examining the transitions in the roles and behaviors of nodes and edges.
To improve the clarity and coherence, our analysis follows a three‐tiered perspectives, beginning with a surface‐level assessment of \textit{performance}, proceeding to an intermediate exploration of the \textit{topology} of knowledge circuits, and culminating in a detailed investigation of the underlying \textit{components}.

\subsection{Performance Analysis}
\label{sec:performance}

An identified knowledge should be capable of independently reproducing the behavioral patterns or performance of the whole model with respect to the corresponding tasks.
This property can be evaluated by examining whether the identified knowledge circuit aligns with the underlying algorithm implemented by the model.
Following \citet{knowledge_circuits}, we employ the Hit@10 metric to measure the rank of the target token among the top 10 predicted tokens throughout training process:
\begin{equation}
    \text { Hit@10 }=\frac{1}{|D_{\text{test}}|} \sum_{i=1}^{|D_{\text{test}}|} \mathrm{I}\left(\operatorname{rank}_a \leq 10\right)
\end{equation}
where $|D_{\text{test}}|$ denotes the test set size, $a$ the target attribute, and $\text{rank}_a$ the rank of the first token of target attribute $a$ in vocabulary space. 
To evaluate completeness, we assess the identified circuit's standalone performance on a held-out test set, which is filtered by the same knowledge type and frequency as the validation set for circuit discovery.

The results depicted in Figure~\ref{fig:hit_at_10} reveal a consistent growth pattern in the Hit@10 metric until it approach its upper bound, which demonstrates the sustained knowledge acquistion capability of knowledge circuits throughout continual pre-training.
Notably, the $K_\text{rel}$ performance curve consistently lies above the curve for $K_\text{compl}$, suggesting that LLMs exhibit preferential learning efficiency when assimilating knowledge extensions within existing conceptual frameworks, as opposed to acquiring completely new knowledge.
These patterns persist in the whole model evaluation in Appendix \ref{app:performance}, suggesting that knowledge circuits capture general learning dynamics rather than isolated phenomena in LLMs.

\begin{tcolorbox}[mybox, title={Takeaway: Knowledge Relevance Principle}]
The acquisition of new knowledge is influenced by its relevance to pre-existing knowledge. LLMs exhibit learning efficiency advantages when acquiring relevant new knowledge versus completely new knowledge.
\end{tcolorbox}


This insight could motivate \textbf{the utilization of data curriculums in continual pre-training}, by organizing the data in a way that mimics the structure and distribution of the original corpus, thereby enabling the model to integrate new information more efficiently \citep{Yıldız_Ravichandran_Punia_Bethge_Ermis_2024,Parmar_Satheesh_Patwary_Shoeybi_Catanzaro_2024,Chen_Chen_Wang_Zhou_Zhu_Jiang_Min_Zhao_Dou_Mao_2024}.

Another notable observation in Figure \ref{fig:hit_at_10} is that the performance of knowledge circuits is positively correlated with knowledge frequency.
We further evaluate the performance of knowledge circuits by transferring them to a test set with different knowledge frequencies, as detailed in Appendix \ref{app:transfer_setting}.
The results imply that the poor performance of knowledge circuits for low-frequency knowledge may stem from insufficient knowledge representations, rather than fundamental capacity limitations of circuits.
This suggests that \textbf{strategies focused on reactivating long-tail knowledge}, such as knowledge augmentation, may improve knowledge retention in LLMs over time \citep{physics3.1}.

\subsection{Topology Analysis}
\label{sec:topology}

In this section, we examine the dynamics of knowledge circuits through a topological lens, employing graph-theoretical metrics to analyze how the circuit subgraphs evolve throughout the training process.

\subsubsection{Structural Consistency}

We first quantify the structural consistency of knowledge circuits by measuring the Jaccard Similarity between edge sets (Figure \ref{fig:similarity_entropy}) and node sets (Figure \ref{fig:nodes_similarity} in Appendix) within knowledge circuits at intermediate checkpoints relative to the final circuit.
Both metrics exhibit a consistent monotonic upward trend throughout training, indicating that the knowledge circuits become increasingly similar to the final circuit.
This convergence pattern suggests an evolutionary process where knowledge circuits progressively stabilize their core architecture as knowledge acquisition progresses.
Based on the observed trends, we hypothesize that the process of knowledge acquisition is driven by topological centralization within knowledge circuits, with a small subset of critical edges and nodes gaining dominance in the flow of information.

\subsubsection{Topological Centralization}

To validate the hypothesis, we define a knowledge circuit entropy metric quantifying edge importance concentration, drawing on the concepts of uncertainty and information content from probability theory and information theory.
The more centralized the topology of the knowledge circuit, the more the importance weights become concentrated on a few critical edges, resulting in a lower knowledge circuit entropy.
To calculate the entropy of a knowledge circuit $\mathcal{C}=<N_{\mathcal{C}},E_{\mathcal{C}}>$, we first normalize the absolute value of the importance of each edge $e\in E_\mathcal{C}$, scored by EAP-IG in equation \eqref{eq:eap-ig}:
\begin{equation}
    P(e)=\frac{S(e)}{\sum_{e^{\prime} \in E_\mathcal{C}} S(e^{\prime})}, \quad \forall e \in E_\mathcal{C}
\end{equation}
The circuit entropy is then calculated as:
\begin{equation}
    H(\mathcal{C})=-\sum_{e\in E_{\mathcal{C}}} P(e)\log P(e)
\end{equation}

\begin{figure*}
    \centering
    \includegraphics[width=\linewidth]{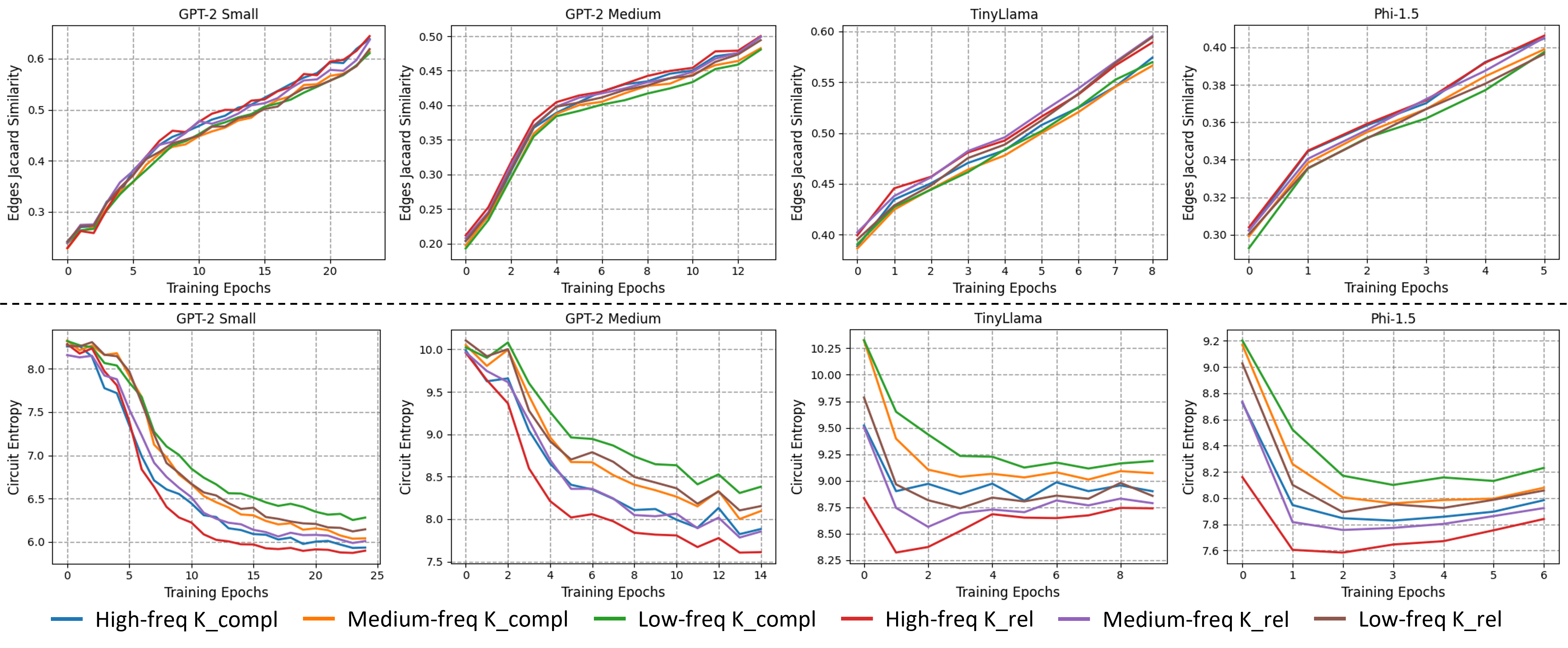}
    \caption{Top: \textbf{Edges Jaccard Similarity} of intermediate knowledge circuits with the circuits at the final checkpoint. Bottom: \textbf{Knowledge Cutcuit Entropy} of knowledge circuits throughout training. \texttt{K\_rel} and \texttt{K\_compl} represent relevant new knowledge and completely new knowledge, respectively. \texttt{Low-freq}, \texttt{Medium-freq}, and \texttt{High-freq} represent knowledge with frequencies in the ranges $[1, 2)$, $[2,5]$ and $(5, 27]$, respectively.}
    \label{fig:similarity_entropy}
\end{figure*}

Our results in Figure~\ref{fig:similarity_entropy} show a stable downward trend in the knowledge circuit entropy metric for edges in the subgraph across all models, suggesting that the identified knowledge circuits become increasingly centralized, with the importance of critical edges growing as knowledge acquisition progresses.
We also observe that the downward trend of the knowledge circuit entropy slows down significantly after a certain turning point during the training of all models.
For example, turning points are observed in GPT-2 Small, GPT-2 Medium, TinyLlama, and Phi-1.5 at epoch 7, epoch 4, epoch 1, and epoch 1, respectively.
We attribute this interesting phenomenon to \textbf{a phase shift in the evolution of knowledge circuits} across continual pre-training.
In the initial \textit{formation phase} of knowledge circuits, less efficient knowledge circuits gradually take shape within the models, resulting in a rapid decrease in circuit entropy.
At the phase shift points, the knowledge circuits reach a status of stability where the most critical nodes and edges have been involved.
In the subsequent \textit{optimization phase}, the topology composed critical nodes and edges becomes more stable, while the computations within these components are being optimized to represent and retrieve the knowledge more efficiently, leading to a slowdown in the rate of decrease in circuit entropy.

It's no coincidence that we also observe consistent phase shift points in the structral consistency of the nodes and edges in knowledge circuits throughout continual pre-training in Figure \ref{fig:similarity_entropy} and Figure \ref{fig:nodes_similarity}, which signal a slowdown in the rate of structural convergence.
This further confirms a reduction in the topological changes of the knowledge circuits, with subsequent performance improvements primarily attributed to the refinement and optimization of the efficiency of the existing structure.

Moreover, we find that the larger the size of the base pre-trained LLMs, the fewer training steps are required to reach the phase shift point in the knowledge circuits evolution.
We suggest that differences in model behavior may stem from the knowledge capacity scaling laws \citep{physics3.3}, which result from a combination of complex factors such as pre-training data signal-to-noise ratio, pre-training duration and model architectures and warrant further investigation in the future.

\begin{tcolorbox}[mybox, title={Takeaway: Biphasic Circuit Evolution}]
The evolution of knowledge circuits exhibits a distinct phase shift from formation to optimization, each marked by unique structural and behavioral characteristics.
\end{tcolorbox}

This finding suggests that \textbf{the state of knowledge circuits could serve as a valuable tracking status for the continual pre-training process}, enabling more informed adjustments to the training method or data in response to different phases.
We leave this potential direction for future research.

\subsubsection{Aligning Topology with Specific Knowledge Circuits}
\label{sec:specific_circuit_performance}

\begin{figure}
    \centering
    \includegraphics[width=\linewidth]{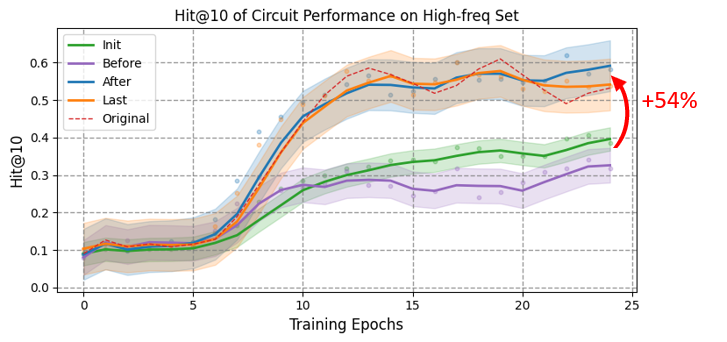}
    \caption{Hit@10 of the performance of aligned knowledge circuits in GPT-2 Small throughout training. \texttt{Init}, \texttt{Before}, \texttt{After},  \texttt{Last} represents the circuits whose topologies align with those at the initial checkpoint, the checkpoint before the phase shift, the checkpoint after the phase shift, and the final checkpoint, respectively. \texttt{Original} represents the original knowledge circuits at each checkpoint. Note that we smooth the curves using a window size of 3 epochs.}
    \label{fig:specific_circuit_performance}
    \vspace{-15pt}
\end{figure}

To clarify the influence of the topology of knowledge circuits on performance, we conduct a detailed examination of the knowledge circuits at several key training checkpoints.
Specifically, we focus on the knowledge circuits at the initial checkpoint, the checkpoint immediately before the phase shift point, the checkpoint immediately after the phase shift point, and the last checkpoint.
We align the topology of the knowledge circuits at each checkpoint throughout training with those of focus and then evaluate the performance for aligned circuits employing the Hit@10 metric as in \S\ref{sec:performance}.
The results in Figure \ref{fig:specific_circuit_performance} reveal that the performance of all aligned circuits remain unchanged during the formation phase.
However, each circuit begins to improve its performance during the optimization phase, with those aligned with the post-phase-shift topologies (\texttt{After} and \texttt{Last}) ultimately performing, on average, 54\% better than those aligned with the pre-phase-shift topologies (\texttt{Init} and \texttt{Before}).
This observation suggests the evolution of the topology of knowledge circuits at the phase shift point plays a crucial role in improving circuit performance.
More examination of the relationship between this topological evolution and the evolution of components will be provided in \S\ref{sec:evolutionary_pattern}.

\subsection{Components Analysis}
\label{sec:components}

After analyzing the dynamics of the knowledge circuits at the overall topology level, we may further seek to understand how the components within these circuits evolve throughout training.

\subsubsection{Evolutionary Pattern of Components}
\label{sec:evolutionary_pattern}

\paragraph{Specialized Nodes}

\begin{figure}
    \centering
    \includegraphics[width=\linewidth]{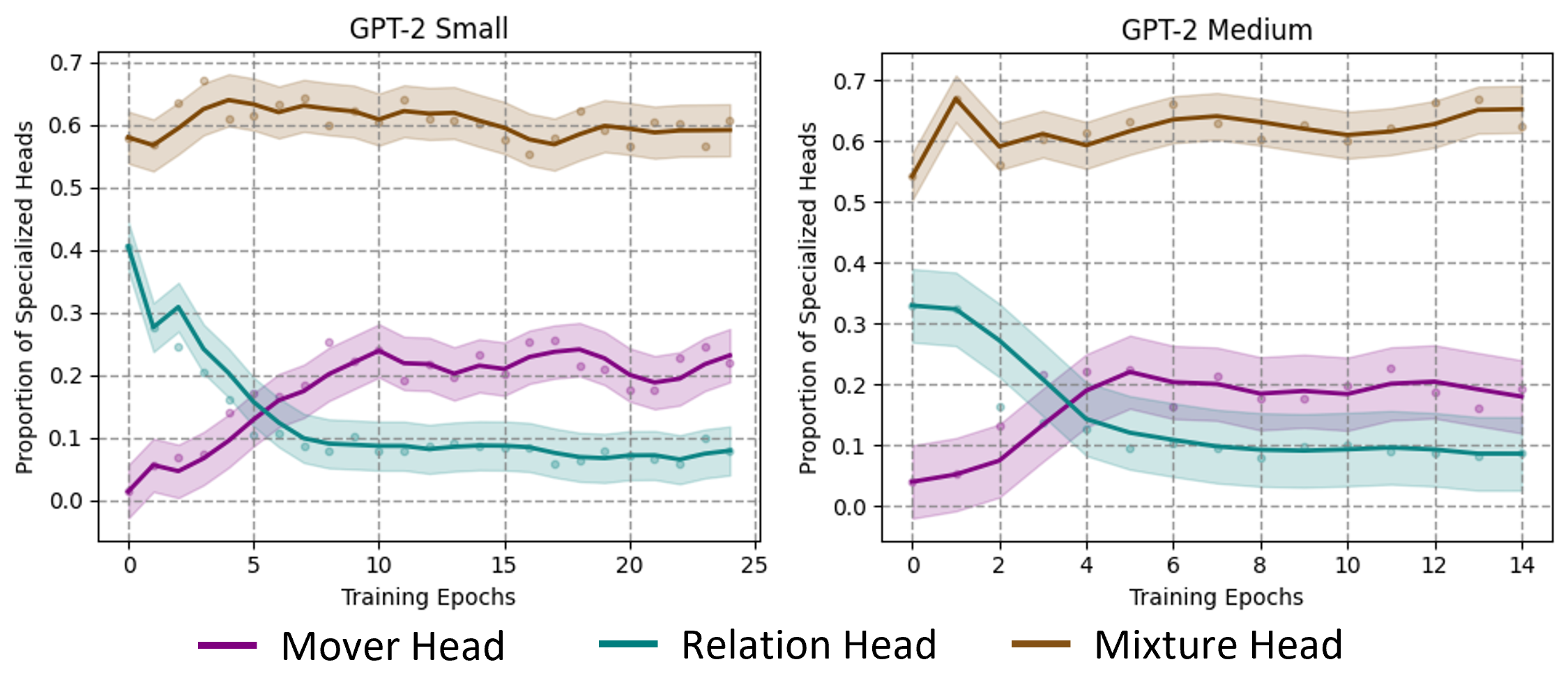}
    \caption{Proportion of \textbf{specialized attention heads} in all nodes of the knowledge circuits throughout training for GPT-2 Small and GPT-2 Medium. Note that we smooth the curves using a window size of 3 epochs.}
    \label{fig:specialized_components}
\end{figure}

We first zoom into the specialized nodes within knowledge circuits to investigate the underlying factors driving the evolution of knowledge circuit.
Recent studies have identified a set of specialized attention heads \citep{attention_heads_survey,mi_primer} that directly contribute to factual recall in Transformer-based LLMs, including the mover head, relation head, and mixture head \citep{DBLP:journals/corr/abs-2403-19521,DBLP:conf/iclr/MerulloEP24,additive_mechanisms}.
More detailed definitions and methodology for identifying these specialized attention heads are provided in Appendix \ref{app:specialized_components}.
We check the emergence and track the proportion of these specialized attention heads in all possible nodes of the knowledge circuits throughout training, and present our results in Figure~\ref{fig:specialized_components}.
We observe that during the circuit formation phase, mover heads gradually emerge from nearly zero, while the proportion of relation heads decreases until the phase shift.
In the circuit optimization phase, the proportion of all kinds of attention heads stabilizes.
The proportion of mixture heads remains stable throughout training.
We further examine the layer-wise distribution of mover heads and relation heads within knowledge circuits throughout training.
Our results in Figure \ref{fig:gpt2_heads_distribution} (and Figure \ref{fig:gpt2_medium_heads_distribution} in Appendix) reveal that the increase in mover heads and the decrease in relation heads primarily occur in the mid-to-deeper layers during the circuit formation phase.

\begin{figure}
    \centering
    \includegraphics[width=\linewidth]{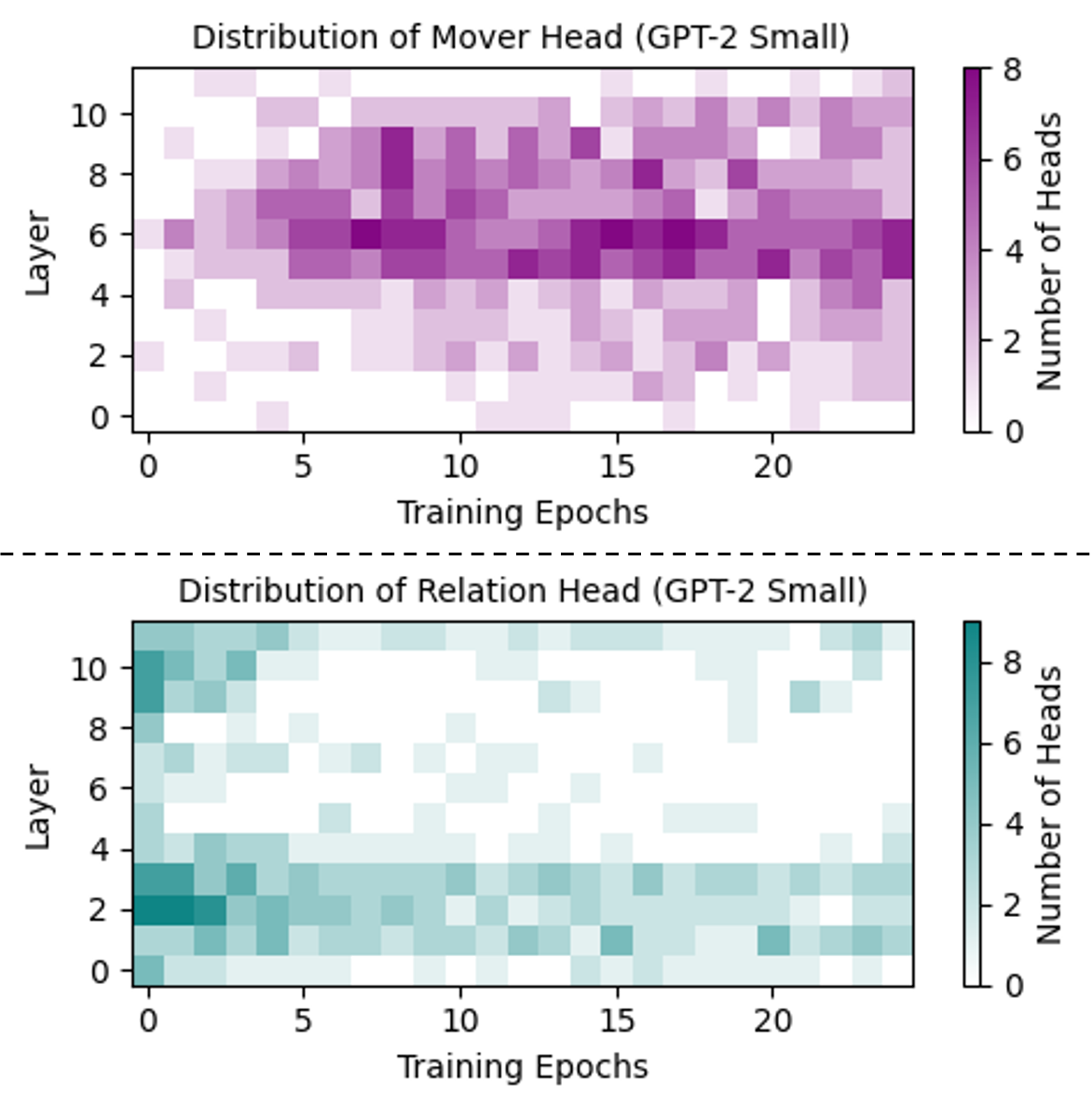}
    \caption{Top: Layer distribution of \textbf{mover head} in the knowledge circuits in GPT-2 Small throughout training. Bottom: Layer distribution of \textbf{relation head} in the knowledge circuits in GPT-2 Small throughout training.}
    \label{fig:gpt2_heads_distribution}
\end{figure}

\paragraph{Activated Edges}

Next, we investigate how the nodes within knowledge circuits propagate information to subsequent components through the edges.
Specifically, we analyze the variation in edge activation patterns across different layers of the network throughout training.
We quantify the edge activation ratio for each layer by calculating the proportion of edges originating from that layer within the knowledge circuit, relative to all possible edges originating from that layer in the whole model\footnote{Note that we exclude the activation ratio for the last layer, as the small denominator causes the ratio to be an outlier, potentially blurring the overall trends in the activation patterns observed across layers.}.
Our results in Figure~\ref{fig:gpt2_activation_ratio} (and Figure \ref{fig:gpt2_medium_activation_ratio} in Appendix) reveal that, during the circuit formation phase, the edges activation ratios in the lower layers gradually decrease, while those in the mid-to-deeper layers exhibit a corresponding increase.
However, as training progresses, a transition occurs around the phase shift point, where the edge activation ratios begin to stabilize.

\begin{figure}
    \centering
    \includegraphics[width=\linewidth]{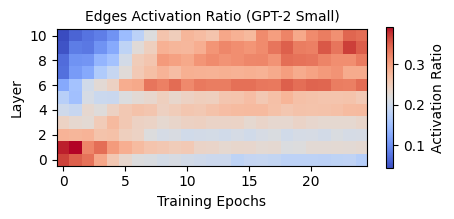}
    \caption{\textbf{Layer distribution of the edges activation ratio} within the knowledge circuits in GPT-2 Small.}
    \label{fig:gpt2_activation_ratio}
\end{figure}

\paragraph{Evolutionary Pattern}
The observed pattern in the evolution of specialized nodes and activated edges within knowledge circuits aligns with the factual recall mechanism in LLMs described by \citet{llm_factual_recall}.
Specifically, the lower MLP layers specialize in encoding attribute-rich subject representations, while attention heads in the mid-to-deeper layers are responsible for extracting the relevant attributes for a given subject from these lower-level representations.
Based on this, we can conclude the evolutionary pattern of knowledge circuits at the component level.
Since we introduce new knowledge entities via synthetic data that the model did not encounter during pre-training, the extraction function is not yet established for these new knowledge entities at the onset of continual pre-training.
Consequently, the model’s attention heads initially concentrate predominantly on the relation tokens already acquired (for example, the city relation learned during pre-training), which manifest as relation heads.
During the early training phase of circuit formation, the focus is primarily on developing the extraction function within the nodes of the mid-to-deeper layers of the knowledge circuits.
With continual pre-training and the gradual acquisition of new knowledge entities, the attention heads in the model’s mid‐to‐upper layers increasingly attended to subject tokens, which were thus classified as mover heads.
This is reflected in the increased emergence of mover heads and activated edges, along with a decrease in the presence of relation heads in these layers.
This process continues until the extraction function is fully established at the phase shift point, as demonstrated by the similar performance advantage of circuits aligned with the post-phase-shift topologies over those aligned with the pre-phase-shift topologies in Figure \ref{fig:specific_circuit_performance}.
In the subsequent training phase of circuit optimization, the focus shifts to enriching knowledge representations in the lower layers, evidenced by a stabilized topology and component structure, but with a rapid improvement in the performance of knowledge circuits in Figure \ref{fig:hit_at_10} and Figure \ref{fig:specific_circuit_performance}.

\begin{tcolorbox}[mybox, title={Takeaway: Deep-to-Shallow Pattern}]
The evolution of knowledge circuits follows a deep-to-shallow pattern, where mid-to-deeper layers first develop the extraction function, and later, lower layers enrich their knowledge representations.
\end{tcolorbox}

\subsubsection{Changes in Vocabulary Space}
\label{sec:rank_prob}

To gain a more nuanced understanding of the information flow, we track the layer-wise changes in both the rank and probability of the target attribute token at the last token position when unembedding the intermediate layer’s output into the vocabulary space throughout training.
Additional results for other models are provided in Appendix \ref{app:rank_prob}.
The results in Figure \ref{fig:gpt2_rank_and_prob} reveal that the occurrence of the early decoding phenomenon \citep{logit_lens}—where the target token is already present in the residual stream by the mid-to-later layers—is closely associated with the phase shift in the evolution of knowledge circuits.
During the circuit formation phase, the mid-to-deeper layers exhibit low ranks and probabilities for the target token, suggesting that the attention heads in these layers have not yet effectively extracted the target attribute in the residual stream due to the insufficient training.
However, in the subsequent circuit optimization phase, the extraction function has already been developed in the mid-to-deeper layers, while the lower layers continue to enrich their knowledge representations for subjects, as evidenced by the occurrence of early decoding phenomenon.

\begin{figure}
    \centering
    \includegraphics[width=\linewidth]{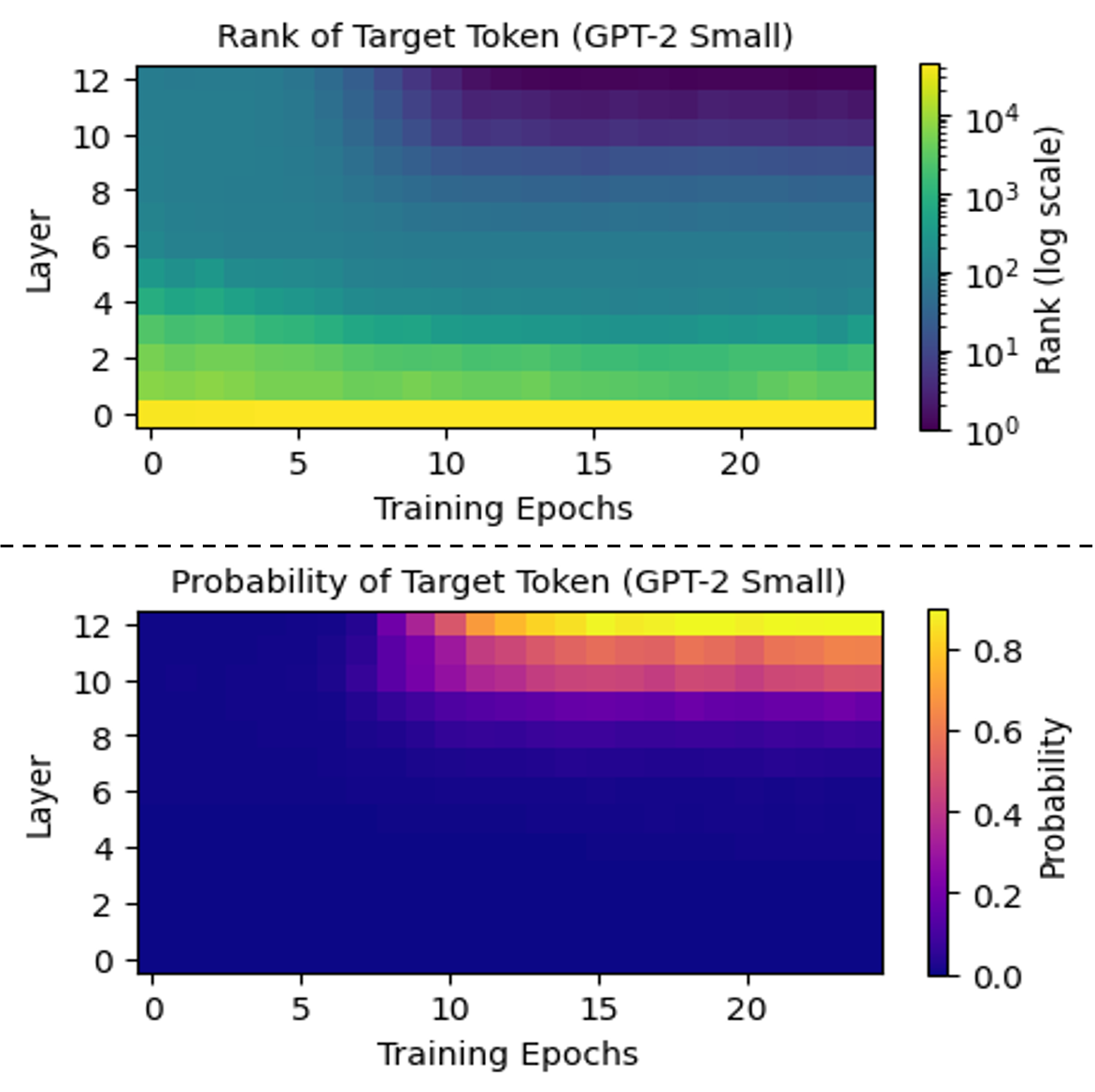}
    \caption{Top: \textbf{Rank of the target attribute token} when unembedding the intermediate layer’s output into vocabulary space at the last token position throughout training for GPT-2 Small. Bottom: The corresponding \textbf{probability of the target attribute token}.}
    \label{fig:gpt2_rank_and_prob}
    \vspace{-10pt}
\end{figure}

\section{Related Work}

\paragraph{New Knowledge Acquisition}

Previous studies~\citep{acquire_factual_knowledge} explore new knowledge acquisition in LLMs with various behavioral interpretability techniques, which characterizes model behavior without revealing insights into the internal workings.
Recent works introduce mechanistic interpretability techniques to advance related research even further.
\citet{physics3.1} adopt probing methods to examine the storage and extraction of factual knowledge encoded in hidden states of language models.
Building on studies that treat feed-forward layers as a key-value memory~\citep{ffn_kv,knowledge_neurons}, \citet{knowledge_entropy} introduce the concept of knowledge entropy to examine how LLMs' knowledge integration evolves during the pre-training phase.
In this paper, we seek to uncover the internal mechanism of new knowledge acquisition in LLMs by investigating the dynamics of knowledge circuits within LLMs thtroughout continual pre-training.

\paragraph{Mechanistic Interpretability}

With the rise of LLMs, Mechanistic Interpretability (MI) has gained prominence for reverse-engineering Transformer-based language models to decode their internal computations~\citep{mi_review,mi_primer,mi_safety_review,singh2024rethinking,sharkey2025open}.
Early MI research identifies features that consistently activate for specific input properties as elementary computational units.
While such studies reveal phenomena such as polysemanticity and enable applications like knowledge editing~\citep{editing_survey,know_edit,DBLP:journals/corr/abs-2406-19354} and steering~\citep{turner2023activation}, they offer limited insights into how features interact to drive model behaviors.
This gap motivates circuit analysis \citep{transformer_circuits,knowledge_circuits}, which investigates computational pathways between Transformer components. 
Most similar to our work, \citet{circuits_are_consistent} examines general circuits formation during pre-training, while our work focuses on the evolution of knowledge circuits throughout continual pre-training.

\section{Conclusion}

In this paper, we present a novel perspective on new knowledge acquisition of LLMs through an investigation into the evolution of knowledge circuits throughout continual pre-training.
Through comprehensive analysis at performance, topology, and components levels, we reveal several key insights.
We believe these insights will contribute to more efficient and effective continual pre-training of LLMs, while also uncovering the mechanisms behind new knowledge acquisition in LLMs.

\section*{Limitations}
\paragraph{Model Architectures}
Our paper investigates the evolution of knowledge circuits solely in decoder-only Transformer LMs, due to their excellent performance and wide range of applications.
We omit other Transformer variants, such as encoder-decoder and encoder-only models, from our analysis.
Additionally, due to limitations in both computational resources and the circuit discovery method, we do not analyze models with larger parameter sizes than 1.3B, which typically employ Grouped Query Attention \citep{GQA}.
However, \citet{circuits_are_consistent} suggests that circuit analyses conducted on small models can provide insights that still apply over model scales.

\paragraph{Traininig Techniques}
We adopt the standard next-token prediction objective for continual pre-training of the base models in our experiments, as it is the most prevalent approach for enabling LLMs to acquire new knowledge.
However, numerous studies~\citep{pit,sft_injecting} focus on designing novel training techniques to enhance the efficiency and effectiveness of LLMs in acquiring new knowledge. 
We do not analyze the impact of these additional training techniques on the evolution of knowledge circuits.

\section*{Acknowledgments}
This work was supported by the National Natural Science Foundation of China (No. 62206246, No. NSFCU23B2055, No. NSFCU19B2027), the Fundamental Research Funds for the Central Universities (226-2023-00138), Yongjiang Talent Introduction Programme (2021A-156-G), CIPSC-SMP-Zhipu Large Model Cross-Disciplinary Fund, Ningbo Natural Science Foundation (2024J020), Information Technology Center and State Key Lab of CAD\&CG, Zhejiang University, the Ministry of Education, Singapore, under the Academic Research Fund Tier 1 (FY2023) (Grant A-8001996-00-00).
We gratefully acknowledge the support of Zhejiang University Education Foundation Qizhen Scholar Foundation.


\bibliography{custom}

\newpage
\appendix
\section*{Appendix}

\section{Dataset Construction}
\label{app:dataset}

Given the challenges of conducting mechanistic interpretability analysis on Internet-scale corpus, we perform controlled experiments on synthetic data, following~\citet{physics3.1,physics3.2,physics3.3}.
We focus on factual knowledge that can be represented as triples of the form $(s, r ,a)$ containing subject $s$, relation $r$, and attribute $a$.
For example, a piece of factual knowledge such as \textit{"Donald Trump is 78 years old"} can be represented as (\textit{Donald Trump, age, 78}).

We synthesize a pool of fictional knowledge entities based on heuristic rules using ChatGPT, ensuring that these fictional biographical knowledge is unavailable to LLMs in the pre-training phase.
Each knowledge entity is first assigned a unique name as the subject.
Each name follows the format \textit{"first\_name middle\_name last\_name"}, where the components are randomly and independently sampled from a uniform distribution.
We use ChatGPT to generate possible values for first name, middle name, and last name,  as listed in Table \ref{tab:names}.

Additionally, there are five associated relations—\textit{birth date}, \textit{city}, \textit{major}, \textit{university} and \textit{company}—which are randomly sampled from their corresponding pools of possible attributes for each relation.
The \textit{birthdate} relation offers 30 (1 to 30) × 12 (January to December) × 126 (1900 to 2025) possibilities.
The corresponding pools of possible attributes for the other four relations are as generated by ChatGPT, as listed in Table \ref{tab:city}$\sim$\ref{tab:university}. 

To convert these entities into textual knowledge for training data, we populate predefined templates with the attribute values.
For each attribute, one of 50 corresponding templates is randomly selected to enhance the diversity of the corpus.
The sentences corresponding to each relation of the same subject are then randomly shuffled to form the biography segment of the subject.
An example is provided below:

\textit{"Liora Shane Driscoll's birth is celebrated annually on \textbf{5 December, 1935}. Liora Shane Driscoll is situated in \textbf{Newport News, VA}. Liora Shane Driscoll is an expert in the making in \textbf{Agronomy}. Liora Shane Driscoll is an alumni member of \textbf{North Carolina State University}. Liora Shane Driscoll is a worker at \textbf{Google}."}

\paragraph{Knowledge Type}
We classify the new knowledge that the language model may need to acquire into two categories. 
One involves knowledge that already exists in the model's parameters but requires further learning of specific aspects (e.g., new relations).
This type of knowledge is referred to as \textit{relevant new knowledge} and denoted as $K_\text{rel}$.
The other type of knowledge is completely new, absent from the model's parameters, which is referred to as \textit{completely new knowledge} and denoted as $K_\text{compl}$.
To simulate real-world data scenarios, we set the knowledge type ratio as $|K_\text{rel}|:|K_\text{compl}|=1:4$.
Specifically, for complete new knowledge, we exclusively use synthetic fictional knowledge entities.
For relevant new knowledge entities, we extract a set of celebrity names from Wikipedia, which are highly likely to appear in pre-training, and then sample fictional attributes for these entities.

\paragraph{Knowledge Frequency}
Considering the long-tail distribution of knowledge in real-world data, we model the frequency of knowledge entities in the corpus to follow an exponential distribution.
This ensures that the corpus for continual pre-training contains both high-frequency knowledge as well as long-tail knowledge.
We classify portions of the corpus based on frequency:
Knowledge entities with a frequency greater than 5 in the corpus are classified as high-frequency knowledge, those with a single occurrence as low-frequency knowledge, and the remaining entities as medium-frequency knowledge.

We set the number of all individuals appearing in the training corpus to 50,000, with their frequency following an exponential distribution between 1 and 27.
This finallly result in 133,408 biography segments, with a total length of 10 million tokens and an average length of 76.8 tokens per biography segment.

\section{Training Configuration}
\label{app:training}
\paragraph{GPT-2}
We adopt the standard GPT-2~\citep{gpt2} implementation available on Huggingface, including GPT-2 Small and GPT-2 Medium.

\paragraph{Llama}
Given the huge experimental cost associated with the original Llama~\citep{llama,llama2,llama3}, which typically have parameters exceeding 7 billion, we perform surrogate experiments using a relatively small model, TinyLlama~\citep{tinyllama}.
TinyLlama adopts exactly the same architecture and tokenizer as Llama 2, but with only 1.1 billion parameters, facilitating more efficient experimentation.

\paragraph{Phi}
We adopt Phi-1.5 \citep{phi-1.5} with 1.3 billion parameters.

For continual-pre training, we use a constant learning rate schedule without warmup.
Our learning rate is set to match the learning rate of the base model at the end of its pre-training phase.
We train using the AdamW optimizer with $\beta_1=0.9,\beta_2=0.95,\epsilon=1e-6$, and a
weight decay of 0.1.
We perform gradient accumulation for every 4 steps.
We present several key statistics of the base models and more hyperparameters that are altered in our experiments in Table \ref{tab:models}.
\begin{table*}[ht]
\renewcommand{\arraystretch}{1.2}
\resizebox{\linewidth}{!}{
\centering
\begin{tabular}{ccccccccc}
\toprule
\multirow{2}{*}{\textbf{Architecture}} & \multirow{2}{*}{\textbf{Model}} & \multicolumn{3}{c}{\textbf{Statistics}} & \multicolumn{4}{c}{\textbf{Hyperparameters}}                \\ \cmidrule(lr){3-5} \cmidrule(lr){6-9}
                        &                         & size     & nodes     & edges    & block\_size & batch\_size & learning\_rate & epochs \\ \midrule
\multirow{2}{*}{GPT-2}  & GPT-2 Small             & 124M         & 158          & 32,491         & 1,024            & 32            &  1e-3              & 25       \\
                        & GPT-2 Medium            & 355M         &  410         &  231,877        & 1,024            & 16            &  1e-3              & 15       \\ \cmidrule(lr){1-9}
Llama               & TinyLlama-v1.1          & 1.1B         &  728         &  742,996        & 2,048            & 4            & 4e-5               & 10       \\ \cmidrule(lr){1-9}
Phi               & Phi-1.5          & 1.3B         &  794  &  886,597        & 2,048            & 2            & 2e-4               & 7       \\\bottomrule
\end{tabular}
}
    \caption{Statistics and hyperparameters of models used in the continual pre-training experiments.}
    \label{tab:models}
    \vspace{-10pt}
\end{table*}

All of our continual pre-training experiments are runned on 2 NVIDIA-A100 GPUs. 

\section{Circuit Discovery}
\label{app:circuit_discovery}

\begin{table}
    \centering
    \resizebox{0.55\linewidth}{!}{
        \begin{tabular}{ll}
        \toprule
            \textbf{Relation} & \textbf{Ratio} \\
            \midrule
            \textit{birthday} & 30 / 30351 \\
            \textit{university} & 102 / 250 \\
            \textit{city} & 151 / 221 \\
            \textit{major} & 138 / 188 \\
            \textit{company} & 142 / 202 \\
        \bottomrule
        \end{tabular}
    }
    \caption{Ratio between the unique first tokens and all the possibilities of the attribute.}
    \label{tab:first_token_ratio}
    \vspace{-10pt}
\end{table}

\begin{table}
    \centering
    \resizebox{0.8\linewidth}{!}{
        \begin{tabular}{ll}
        \toprule
            \textbf{Relation} & \textbf{Template} \\
            \midrule
            \textit{city} & \textit{s} lives in the city of \\
            \textit{major} & \textit{s} majors in the field of \\
            \textit{company} & \textit{s} works for the company of \\
        \bottomrule
        \end{tabular}
    }
    \caption{Templates for the factual recall task on relations.}
    \label{tab:templates}
    \vspace{-10pt}
\end{table}

\subsection{Tasks}
Unlike previous works that investigate circuits on simple but general tasks such as Indirect Object Identification (IOI) and Greater-Than, our paper focuses on knowledge circuits that are capable of performing the task of factual recall.
In a factual recall task, the objective is to predict a target attribute $a$ given a subject-relation pair $(s, r)$.
To ensure a sufficiently rich vocabulary space for the first token of the target attribute, we construct the factual recall tasks based on three relations mentioned in \S\ref{sec:dataset}: \textit{city}, \textit{major}, and \textit{company}.
We exclude the attributes \textit{birthday}, whose first token is always an Arabic numeral between 1 and 30, and \textit{university}, whose first token is typically “University,” from our analysis.
We further supplement Table~\ref{tab:first_token_ratio} by computing the ratio of unique first tokens to the total number of possible values for each attribute.
The findings reveal that the proportion of generic first subtokens is low (approximately 30 \%) for the remaining three attributes \textit{city}, \textit{major}, and \textit{company}, thereby mirroring real‐world distributions without materially affecting performance evaluation.

The templates for converting a subject-relation pair $(s, r)$ into a query string for each factual recall task are listed in Table \ref{tab:templates}.
A typical circuits task consists of minimal pairs of clean and corrupted inputs.
For clean inputs, we randomly sample 300 examples from the training corpus for each knowledge type and frequency as the validation set $D_{\text{val}}$ for circuit discovery.
In our experiments, we observe that continually increasing the size of $D_{\text{val}}$ only adds to the runtime for circuit discovery without improving the quality of the discovered circuits.
The corresponding corrupted inputs are independently sampled from the training corpus to match the length of the subject tokens in each clean input.

\subsection{Loss}
The metric for circuit tasks assesses how closely the language model outputs align with clean input, as opposed to corrupted input.
In our circuit discovery experiments, we evaluate the performance of circuits using the logit difference: the logit of the correct attribute minus the logit of the corrupted attribute.
We then convert the task metric $M$ into a loss function by defining $L(x)=-M(x)$, as shown in Eq.~\ref{eq:eap-ig}.

We make modifications to the TransformerLens library \citep{nanda2022transformerlens} and EAP-IG library \citep{eap-ig} to implement the circuit dicovery method and conduct all the analysis experiments.

\section{Whole Model Performance}
\label{app:performance}

\begin{figure*}
    \centering
    \includegraphics[width=\linewidth]{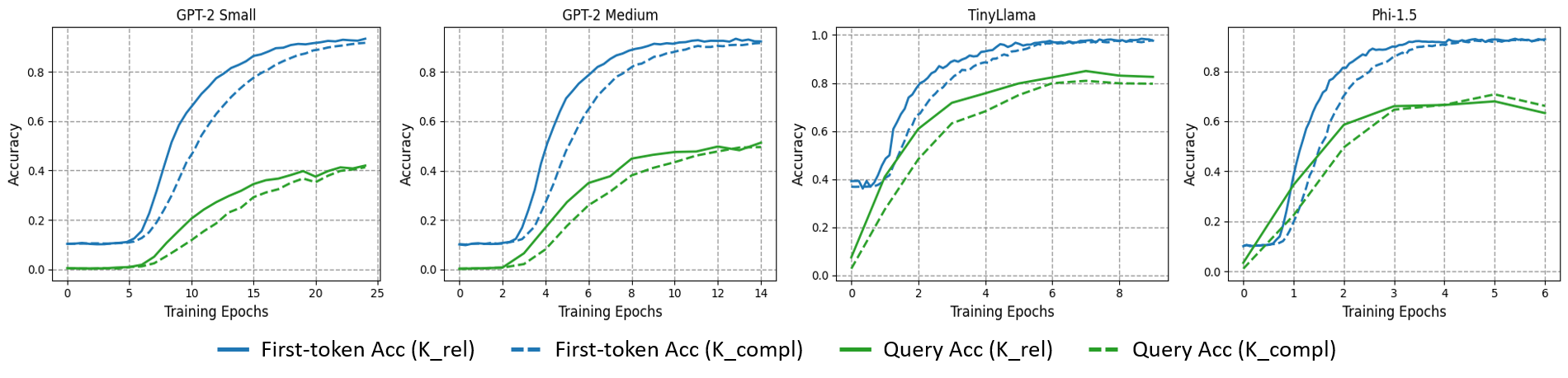}
    \caption{Accuracy curves across continual pre-training. \texttt{K\_rel} and \texttt{K\_compl} represent relevant new knowledge and completely new knowledge, respectively. \texttt{First-token Acc} stands for the model's next-token prediction accuracy on the first token of each attribute, while \texttt{Query Acc} stands for the generation accuracy on downstream query tasks for each attribute.}
    \label{fig:accuracy}
\end{figure*}

We examine the whole model's performance for knowledge acquisition by monitoring two type of accurracies during training process.
First, we track the model's next-token prediction accuracy on the first token of each attribute during training.
This metric reflects how well the model acquires and memorizes the knowledge.
The second metric is calculated on downstream query tasks in cloze-style for each attribute, such as \textit{"s lives in the city of \_\_\_"}, where the accuracy reflects the model's ability to generate an exact match for the correct attribute.
Our results in Figure~\ref{fig:accuracy} illustrate that both accuracy metrics increase until they reach their upper limits, reflecting the model's ongoing acquisition of new knowledge during continual pre-training.
Another interesting observation is that the accuracy curve for $K_\text{rel}$ consistently lies above the curve for $K_\text{compl}$ on both metrics, suggesting that relevant new knowledge is easier for LLMs to acquire than completely new knowledge.

\section{Transfer Performance of Knowledge Circuits between Frequency}
\label{app:transfer_setting}
To investigate the differences in the capacities of knowledge circuits identified using validation data filtered by knowledge frequency, we analyze the transfer performance of these circuits on held-out test sets with varying transferred knowledge frequencies.
For example, if a knowledge circuit is identified using validation data filtered by high-frequency knowledge, denoted as \texttt{High-freq Circuit}, its transfer performance is evaluated on test sets filtered by medium-frequency and low-frequency knowledge, respectively.

\begin{figure*}
    \centering
    \includegraphics[width=\linewidth]{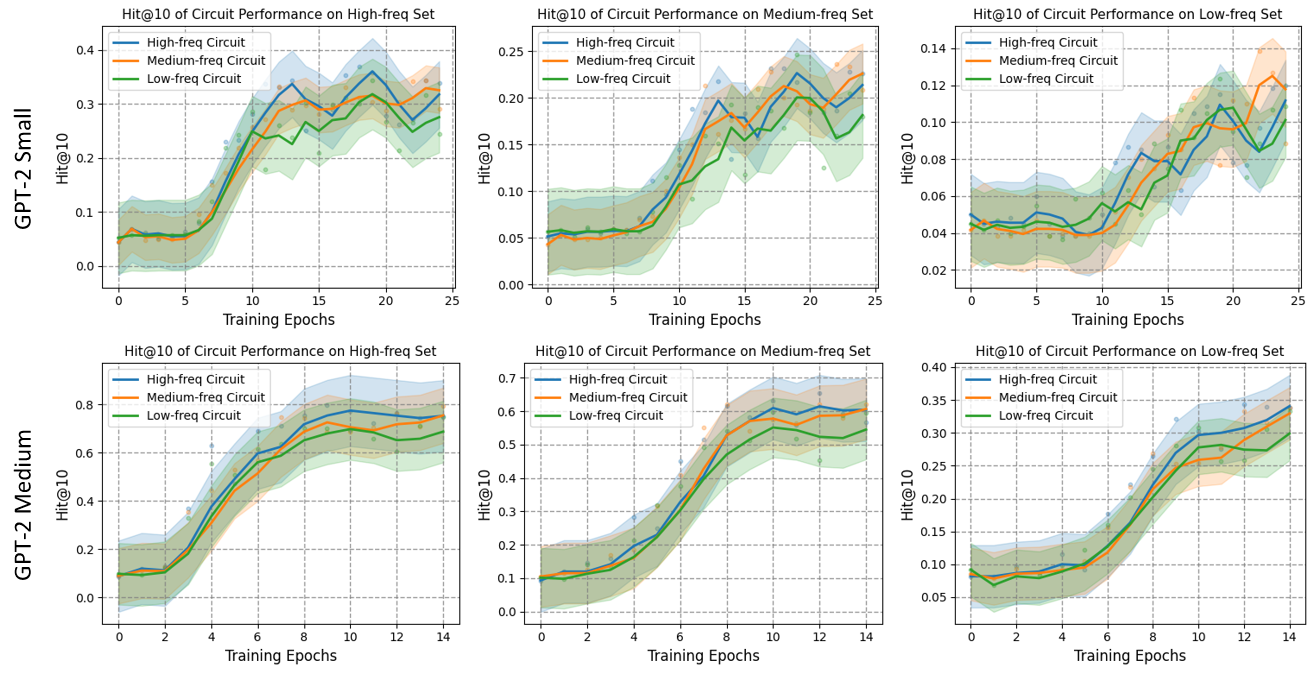}
    \caption{Hit@10 of the transfer performance of knowledge circuits in GPT-2 Small and GPT-2 Medium throughout training. \textcolor[RGB]{44,160,44}{\texttt{Low-freq Circuit}}, \textcolor[RGB]{255,127,14}{\texttt{Medium-freq Circuit}}, and \textcolor[RGB]{31,119,180}{\texttt{High-freq Circuit}} represent knowledge circuits identified by knowledge with the frequencies in the ranges  \textcolor[RGB]{44,160,44}{$[1, 2)$}, \textcolor[RGB]{255,127,14}{$[2,5]$} and \textcolor[RGB]{31,119,180}{$(5, 27]$}, respectively. Note that we smooth the curves using a window size of 3 epochs for all settings.}
    \label{fig:transfer_hit_at_10}
\end{figure*}

Our results in Figure \ref{fig:transfer_hit_at_10} reveal that knowledge circuits identified using knowledge of different frequencies perform comparably when evaluated on test sets of the same frequency. Notably, knowledge circuits discovered using high-frequency knowledge exhibit relatively poor performance on the low-frequency test set, whereas circuits identified using low-frequency knowledge perform comparably to high-frequency circuits on the high-frequency test set. This finding suggests that there is no inherent difference in the capability of circuits for the same task; rather, their effectiveness is primarily determined by the representation of knowledge shaped by frequency.

\begin{figure*}
    \centering
    \includegraphics[width=\linewidth]{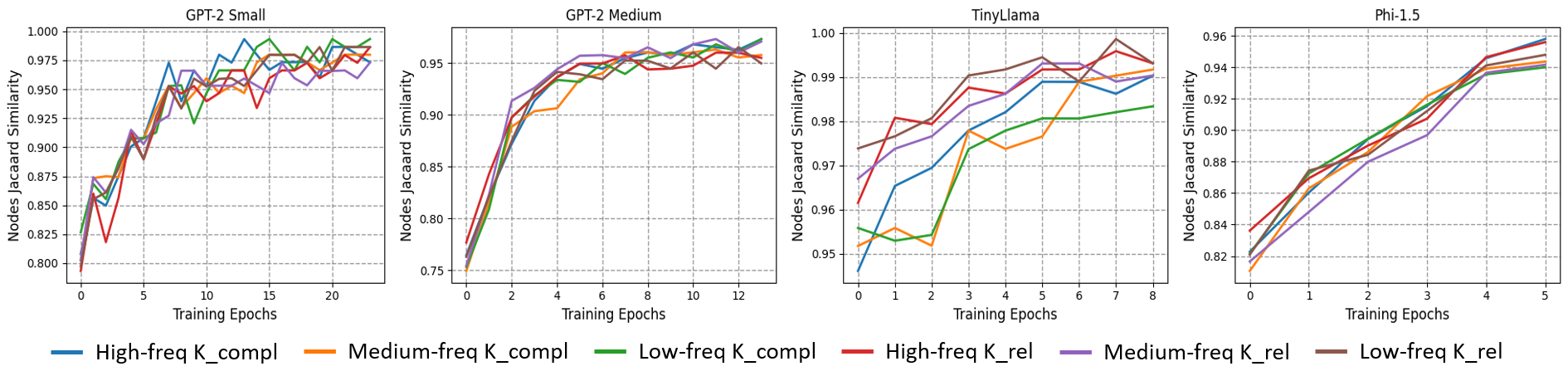}
    \caption{\textbf{Nodes Jaccard Similarity} of intermediate knowledge circuits with the circuits at the final checkpoint. \texttt{K\_rel} and \texttt{K\_compl} represent relevant new knowledge and completely new knowledge, respectively. \texttt{Low-freq}, \texttt{Medium-freq}, and \texttt{High-freq} represent knowledge with frequencies in the ranges $[1, 2)$, $[2,5]$ and $(5, 27]$, respectively.}
    \label{fig:nodes_similarity}
\end{figure*}

\begin{figure}
    \centering
    \includegraphics[width=0.8\linewidth]{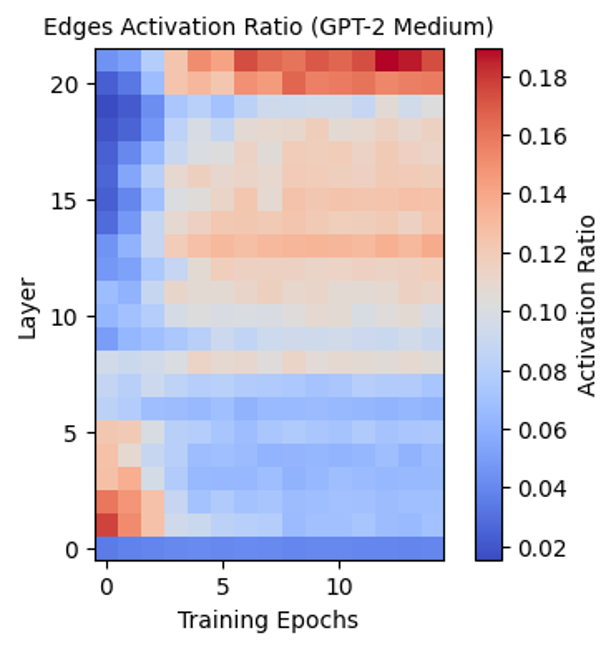}
    \caption{\textbf{Layer distribution of the edges activation ratio} within the knowledge circuits in GPT-2 Medium.}
    \label{fig:gpt2_medium_activation_ratio}
    \vspace{-10pt}
\end{figure}

\section{Changes in Vocabulary Space}
\label{app:rank_prob}

\begin{figure*}
    \centering
    \includegraphics[width=\linewidth]{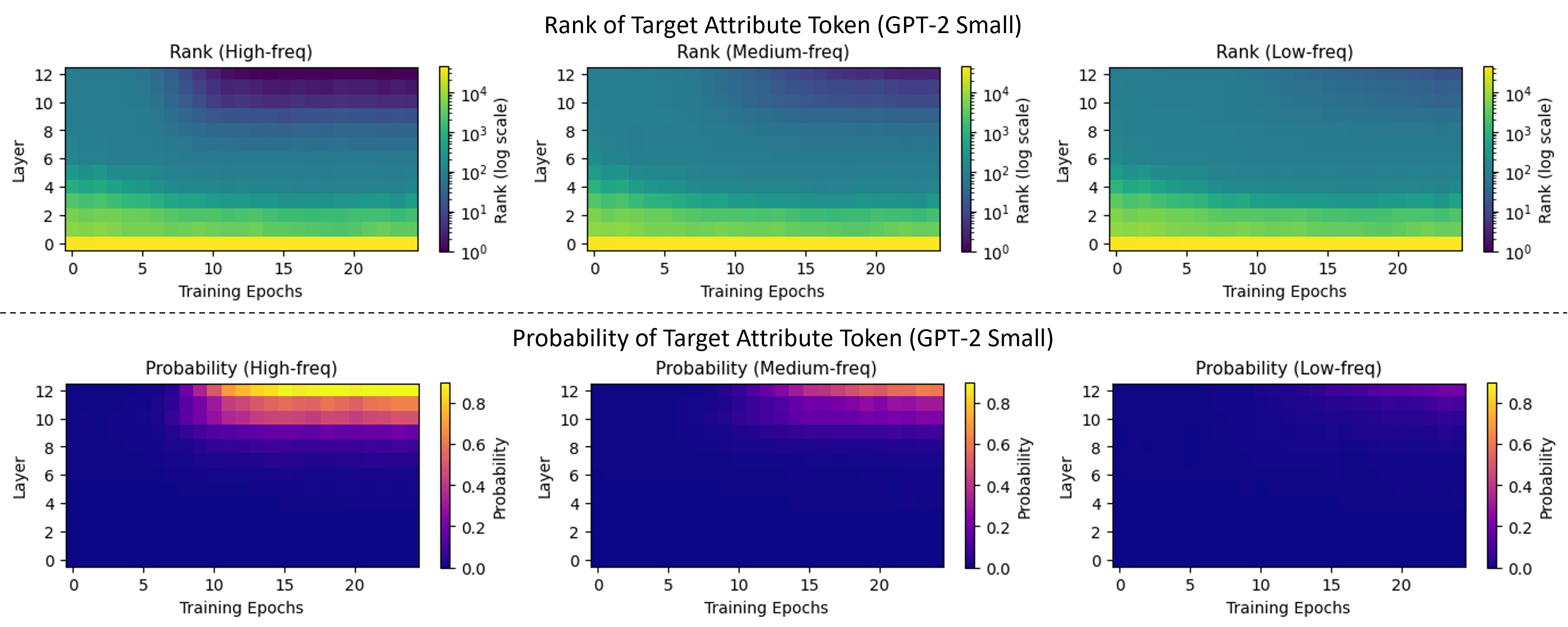}
    \caption{Top: \textbf{Rank of the target attribute token} when unembedding the intermediate layer’s output into vocabulary space at the last token position throughout training for GPT-2 Small. Bottom: \textbf{Probability of the target attribute token} when unembedding the intermediate layer’s output into vocabulary space at the last token position throughout training for GPT-2 Small. \texttt{Low-freq}, \texttt{Medium-freq}, and \texttt{High-freq} represent knowledge with frequencies in the ranges $[1, 2)$, $[2,5]$ and $(5, 27]$, respectively.}
    \label{fig:gpt2_all_rank_and_prob}
    \vspace{-10pt}
\end{figure*}

\begin{figure*}
    \centering
    \includegraphics[width=\linewidth]{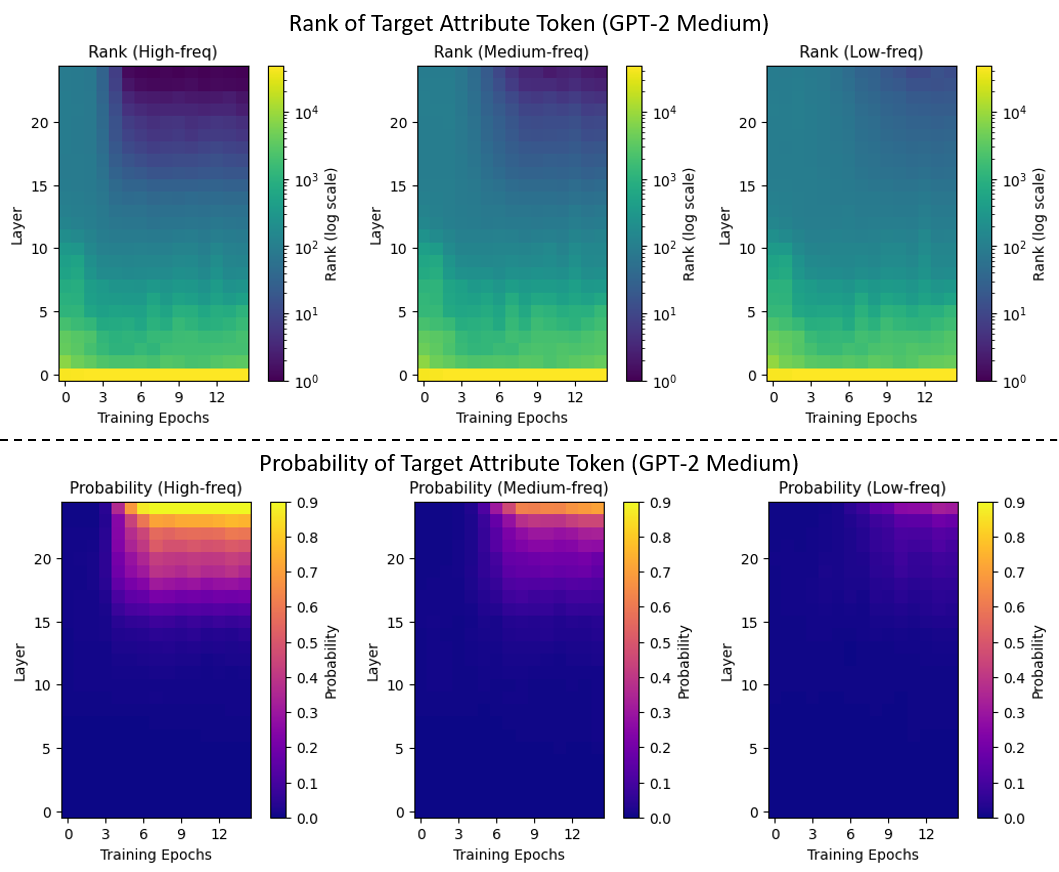}
    \caption{Top: \textbf{Rank of the target attribute token} when unembedding the intermediate layer’s output into vocabulary space at the last token position throughout training for GPT-2 Medium. Bottom: \textbf{Probability of the target attribute token} when unembedding the intermediate layer’s output into vocabulary space at the last token position throughout training for GPT-2 Medium. \texttt{Low-freq}, \texttt{Medium-freq}, and \texttt{High-freq} represent knowledge with frequencies in the ranges $[1, 2)$, $[2,5]$ and $(5, 27]$, respectively.}
    \label{fig:gpt2_medium_rank_and_prob}
    \vspace{-10pt}
\end{figure*}

\begin{figure*}
    \centering
    \includegraphics[width=\linewidth]{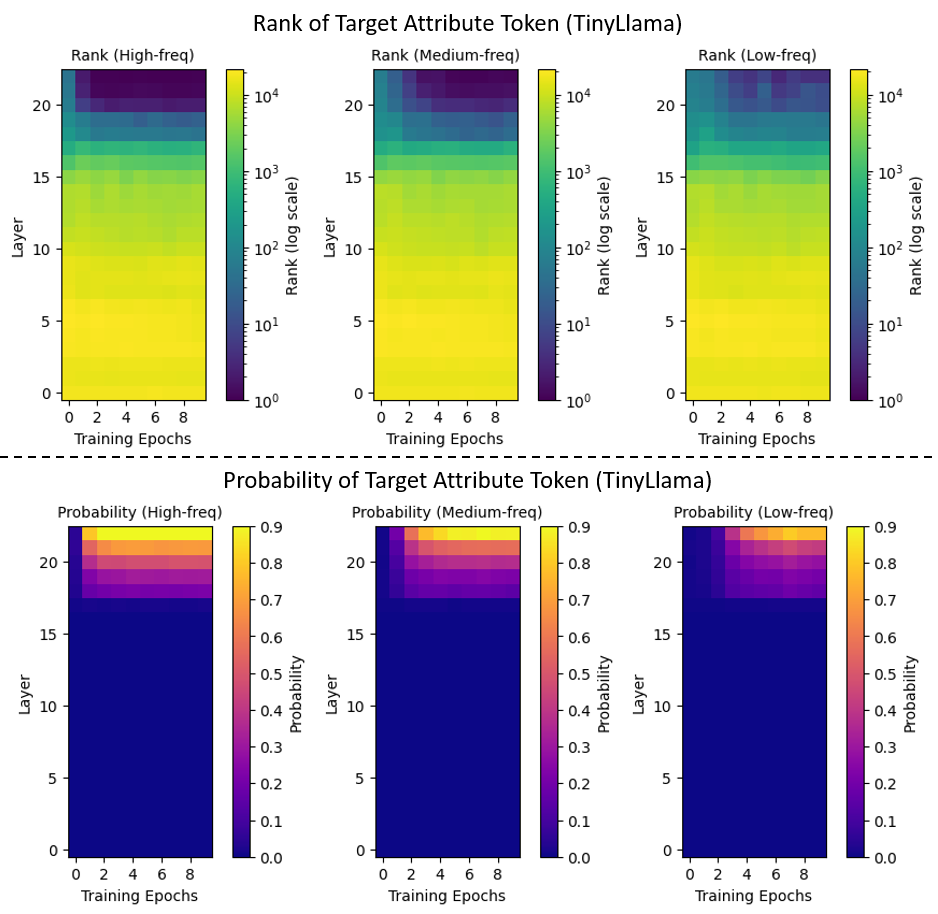}
    \caption{Top: \textbf{Rank of the target attribute token} when unembedding the intermediate layer’s output into vocabulary space at the last token position throughout training for TinyLlama. Bottom: \textbf{Probability of the target attribute token} when unembedding the intermediate layer’s output into vocabulary space at the last token position throughout training for TinyLlama. \texttt{Low-freq}, \texttt{Medium-freq}, and \texttt{High-freq} represent knowledge with frequencies in the ranges $[1, 2)$, $[2,5]$ and $(5, 27]$, respectively.}
    \label{fig:tinyllama_rank_and_prob}
    \vspace{-10pt}
\end{figure*}

We present our results for the layer-wise changes in the rank and probability of the target attribute token at the final token position when unembedding the intermediate layer’s output into the vocabulary space throughout the training of GPT-2 Small on the high-frequency set in \S\ref{sec:rank_prob}.
Additionally, we provide the full results of all knowledge frequencies for GPT-2 Small in Figure \ref{fig:gpt2_all_rank_and_prob}.
We also provide the full results for GPT-2 Medium (Figure \ref{fig:gpt2_medium_rank_and_prob}) and TinyLlama (Figure \ref{fig:tinyllama_rank_and_prob}).

\section{Specialized Attention Heads within Knowledge Circuits}
\label{app:specialized_components}

\subsection{Definitions of Specialized Attention Heads}

When zooming into the discovered knowledge circuits, we can find several specialized attention heads in the model that play a crucial role in the final prediction.
These include the mover head, relation head, and mixture head \citep{additive_mechanisms}.

\paragraph{Mover head}
Attention head that focuses on the final token of the context and attends strongly to the subject tokens in the context, functioning as a mover to transfer information and extract attributes pertaining to the subject from the enriched subject representation.

\paragraph{Relation head}
Attention head that focuses on the final token of the context and attends strongly to the relation tokens in the context for a particular relation and extract many relation-related attribute tokens.

\paragraph{Mixture head}
Attention Head that attends to both the relation tokens and the subject tokens in the context.
It behaves as a combination of the two, performing the role of both Mover Head and Relation Head simultaneously.

\subsection{Identification of Specialized Attention Heads}
In this section, we provide details on how to identify mover heads, relation heads, and mixture heads in LLMs.
We re-implement the methodology described in \citet{additive_mechanisms} since the original code has not been made publicly available by the authors.
We will update our implementation once the source code is released.

Building on the Direct Logit Attribution (DLA) technique, which measures the direct effect of individual model components on model outputs, \citet{additive_mechanisms} move beyond and propose DLA by source token group.
This technique is based on the observation that attention head outputs are a weighted sum of outputs corresponding to distinct attention source positions \citep{transformer_circuits}.
This approach is useful for quantifying how a source token group directly affects the logits through individual attention heads.

With a specific factual recall task where the relation held constant, we aggregate over the validation set $D_{\text{val}}$ for circuit discovery on the task, an attention head is classified as mover head if:
\begin{equation}
   \left|\frac{\sum_{i=1}^{|D_{\text{val}}|}\text{DLA}_s(\text{Q}_i)}{\sum_{i=1}^{|D_{\text{val}}|}\text{DLA}_r(\text{Q}_i)}\right| > \tau
\end{equation}
where $i$ denotes the $i$-th entity in $D_{\text{val}}$, $\text{Q}_i$ denotes the relation-specific query string for entity $i$ as shown in Table \ref{tab:templates}, $\text{DLA}_s(\text{Q}_i)$ denotes DLA attributed to subject tokens, and $\text{DLA}_r(\text{Q}_i)$ denotes DLA attributed to relation tokens.
Relatively, an attention head is classified as relation head if:
\begin{equation}
    \left|\frac{\sum_{i=1}^{|D_{\text{val}}|}\text{DLA}_s(\text{Q}_i)}{\sum_{i=1}^{|D_{\text{val}}|}\text{DLA}_r(\text{Q}_i)}\right| < \frac{1}{\tau}
\end{equation}
where threshold $\tau$ is set to be 10 as suggested in \citet{additive_mechanisms}.
Remaining attention heads in LLMs are classified as mixture heads, behaving as a combination of mover head and relation head.

\begin{figure}
    \centering
    \includegraphics[width=\linewidth]{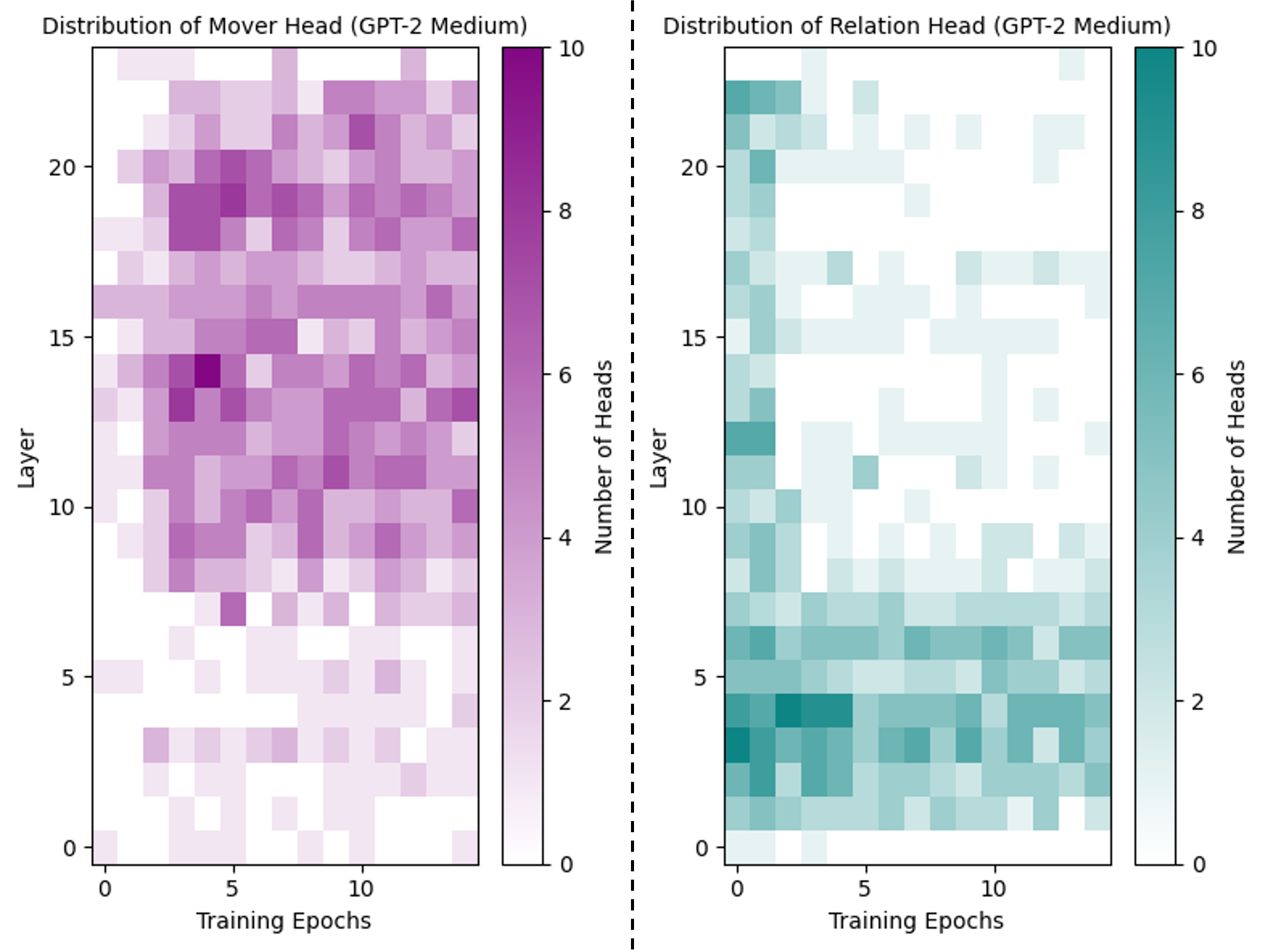}
    \caption{Left: Layer distribution of \textbf{mover head} in the knowledge circuits in GPT-2 Medium throughout training. Right: Layer distribution of \textbf{relation head} in the knowledge circuits in GPT-2 Medium throughout training.}
    \label{fig:gpt2_medium_heads_distribution}
    \vspace{-15pt}
\end{figure}

\section{Forgetting Analysis for Knowledge Circuits}

\begin{figure}
    \centering
    \includegraphics[width=\linewidth]{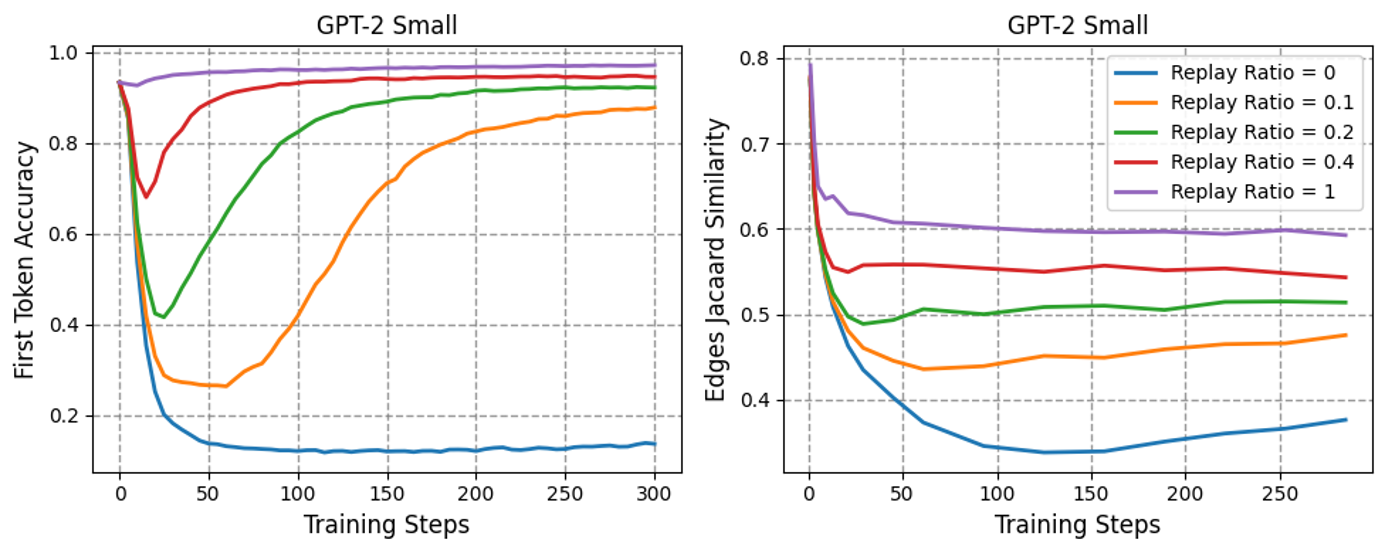}
    \caption{Edges Jaccard Similarity of intermediate knowledge circuits with the circuits at the final checkpoint of the previous knowledge acquisition experiment.}
    \label{fig:gpt2_forget}
    \vspace{-10pt}
\end{figure}



To analyze the model's forgetting of acquired knowledge, we conduct and additinal coninual pre-training experiment.
We first construct new training corpus following the same pipeline described in \S\ref{sec:dataset}, and then initialize training from the final checkpoint of the previous knowledge acquisition experiment on GPT-2 Small.
We monitor structral consistency changes for knowledge circuits throughout 10 training epochs.

Our results in Figure \ref{fig:gpt2_forget} reveal that knowledge circuits demonstrate structural reconfiguration capacity, with the identified circuits dynamically adjusting more than 60\% of their edges to accommodate new knowledge.
However, data replay interventions, which involve the periodic replacement of a fixed ratio of original training samples, successfully mitigate knowledge forgetting by reactivating circuit components.
This evidence suggests that LLMs maintain latent reactivation potential even after apparent behavioral forgetting — a property we term knoweldge circuit elasticity.


\begin{table*}
    \centering
    \resizebox{\textwidth}{!}{
        \begin{tabular}{lp{21cm}}
        \hline
            \textbf{Name} & \textbf{Possible Values} \\ \hline
            First Name & Aarav, Abbott, Aberdeen, Abilene, Acey, Adair, Adelia, Adriel, Afton, Aida, Ainsley, Aislinn, Alaric, Albin, Alden, Aleah, Alessandra, Alistair, Allegra, Alphonse, Althea, Amaury, Ambrose, Amelina, Amias, Anatole, Anders, Ansel, Anthea, Antonella, Anwen, Arden, Ariadne, Aric, Arlen, Armand, Armando, Arwen, Asa, Astra, Atticus, Aubrey, Auden, Aurelia, Aurora, Aveline, Aviana, Azariah, Baird, Basil, Bayard, Beauregard, Bellamy, Belvedere, Benedict, Bennett, Berenice, Bertram, Blaine, Blair, Blythe, Boaz, Bodhi, Boniface, Bram, Branwen, Brenna, Briar, Briony, Broderick, Bromley, Bronson, Cadence, Cael, Caelan, Caius, Caledon, Calista, Calliope, Callum, Calyx, Cambria, Camellia, Candela, Caspian, Cassian, Cassiopeia, Castor, Cecily, Celeste, Celestia, Cerelia, Cerys, Chalcedony, Chandra, Charlton, Cicero, Cillian, Clemence, Clementine, Cleo, Clio, Clovis, Colton, Conall, Conrad, Corbin, Cordelia, Cormac, Cosima, Cressida, Crispin, Cybele, Cyril, Dahlia, Damaris, Daphne, Darby, Darcy, Dario, Davina, Deirdre, Delaney, Delphine, Demelza, Desmond, Dexter, Dimitri, Dinah, Dorian, Dulcie, Eamon, Earlene, Eben, Edeline, Edmund, Eldon, Eleri, Elia, Elian, Elias, Elodie, Eloise, Elowen, Ember, Emeline, Emrys, Endellion, Ender, Ephraim, Erasmus, Esme, Eulalia, Evadne, Evander, Everard, Everett, Fable, Fanchon, Farrah, Faye, Felix, Fern, Finlay, Fiora, Fletcher, Florian, Forsythia, Freya, Frida, Gable, Galen, Gareth, Garnet, Garrick, Gelsey, Gemma, Genever, Genevieve, Ginevra, Grady, Griffin, Guinevere, Hadley, Halcyon, Hale, Harlan, Hart, Haven, Hawthorne, Hazel, Heath, Helena, Hesper, Hollis, Honora, Hyacinth, Idris, Ilaria, Ilona, Imara, Indigo, Ingrid, Ione, Iris, Isadora, Isolde, Ivor, Jago, Jareth, Jarvis, Jemima, Jericho, Jocasta, Jolyon, Jorah, Jory, Jovan, Jubilee, Jules, Junia, Juniper, Kael, Kaia, Kalista, Kalliope, Katriel, Keir, Kenna, Kerensa, Keturah, Keziah, Kieran, Kirby, Kismet, Kit, Knox, Kyrie, Lachlan, Lark, Larkin, Laszlo, Leda, Leif, Lennox, Leonie, Leopold, Leta, Linnea, Liora, Livia, Llewellyn, Locke, Lorcan, Lorelei, Lorna, Lucian, Lysandra, Lysander, Mabel, Macey, Maeve, Magnolia, Malachi, Malin, Manon, Marcel, Marcellus, Maren, Marius, Marisol, Maris, Mathis, Matilda, Mavis, Maximilian, Meadow, Merrick, Merritt, Micaiah, Micah, Mira, Mireille, Mireya, Mirren, Morrigan, Muir, Nadia, Nadine, Nairne, Nara, Nash, Navi, Naylor, Neve, Nico, Nina, Noble, Nolan, Nora, Nova, Nyssa, Oberon, Octavia, Odessa, Oisin, Oleander, Olwen, Onyx, Ophelia, Orion, Orla, Orson, Osiris, Osric, Ottilie, Ozias, Paisley, Paloma, Pax, Paz, Penelope, Peregrine, Persephone, Phaedra, Phineas, Phoenix, Pippa, Poppy, Portia, Posy, Primrose, Quill, Quinlan, Rafferty, Rain, Rainer, Raphael, Raven, Reeve, Reinette, Renata, Rhea, Rhiannon, Rhys, Riona, Roderick, Romilly, Rowan, Roxana, Rufus, Sable, Sabine, Saffron, Sage, Salem, Samara, Sancia, Saoirse, Sarai, Saskia, Selah, Seneca, Seraphina, Seren, Severin, Shai, Shiloh, Sibyl, Sidonie, Silas, Simeon, Simone, Sinclair, Sol, Solange, Sorrel, Sparrow, Stellan, Sullivan, Sylvain, Sybil, Sylvana, Tallulah, Tamsin, Tansy, Tarquin, Taryn, Tavish, Tegan, Thaddeus, Thelma, Theodora, Theron, Thorin, Thorne, Thora, Tiernan, Tristan, Tullia, Ursula, Valencia, Valerian, Vanya, Vesper, Vianne, Violetta, Virgil, Waverly, Wendell, Willa, Windsor, Winston, Wisteria, Wren, Wynn, Xanthe, Xavier, Xenia, Xerxes, Yara, Yasmin, Yelena, Ysabel, Yvaine, Zahra, Zara, Zephyr, Zinnia, Ziva, Zora
            \\ \hline
            Middle Name &  Abel, Abram, Ace, Adele, Ainsley, Alaric, Alcott, Alden, Allegra, Amara, Amethyst, Anders, Ansel, Arden, Arlo, Arrow, Asa, Asher, Aster, Astrid, Atticus, Auden, Aurora, Austen, Axel, Baird, Basil, Bay, Beau, Beck, Blaise, Blake, Blythe, Boden, Bodhi, Boone, Bram, Bran, Briar, Briggs, Brooks, Calla, Calvin, Caspian, Cassian, Cedar, Celeste, Chance, Channing, Cleo, Clove, Clyde, Cohen, Colt, Cove, Crew, Crosby, Cyrus, Dane, Dante, Dashiell, Dawn, Dax, Dean, Delta, Dimitri, Dove, Drake, Dune, Echo, Eden, Edison, Elara, Elian, Ellis, Elowen, Ember, Emrys, Eos, Esme, Evangeline, Ever, Everest, Ewan, Eyre, Fable, Fairfax, Fallon, Faye, Fenton, Fern, Finnian, Fleur, Flynn, Forrest, Fox, Gage, Gale, Garnet, Gideon, Gray, Greer, Halcyon, Hale, Harlow, Haven, Hawk, Hayes, Hollis, Hope, Hugo, Idris, Iker, Indigo, Ines, Iona, Iris, Isla, Iver, Jace, Jade, Jagger, Jem, Jet, Joaquin, Jude, Jules, Kai, Kane, Kash, Keats, Keira, Kellen, Kendrick, Kepler, Kian, Kit, Knox, Lake, Lark, Laurel, Layne, Leif, Lennox, Lester, Levi, Liam, Lila, Linnea, Locke, Lorcan, Lore, Luca, Lucian, Lux, Lyric, Maeve, Magnus, Maia, Malcolm, March, Maren, Marlow, Mason, Maverick, Meadow, Mercer, Merrick, Mica, Milan, Milo, Monroe, Moon, Nash, Nico, Noble, Noor, North, Oak, Oberon, Odette, Oisin, Oleander, Onyx, Opal, Orion, Otis, Otto, Pace, Parker, Pax, Paz, Penn, Perry, Phoenix, Pierce, Pine, Poe, Poet, Poppy, Porter, Prosper, Quill, Quincy, Rain, Reed, Reeve, Remy, Rex, Rhea, Ridge, Riven, Roan, Rogue, Roman, Rook, Rowan, Rune, Sable, Sage, Sailor, Saxon, Scout, Sequoia, Shane, Shiloh, Sierra, Sloane, Sol, Solstice, Soren, Sparrow, Star, Stone, Storm, Story, Sullivan, Sylvan, Talon, Tamsin, Tate, Teague, Teal, Thane, Thatcher, Thorn, Thornton, Tide, Torin, True, Vail, Valor, Veda, Vesper, Vince, Violette, Wade, Waverly, Wells, West, Wilder, Willow, Winter, Wren, Wynn, Xander, Xanthe, Xavier, Yara, York, Yule, Zane, Zara, Zephyr, Zinnia
            \\ \hline
            Last Name & Abernathy, Ainsworth, Alberts, Ashcroft, Atwater, Babcock, Bader, Bagley, Bainbridge, Balfour, Barkley, Barlowe, Barnhill, Biddle, Billingsley, Birkett, Blakemore, Bleeker, Bliss, Bonham, Boswell, Braddock, Braithwaite, Briggs, Brockman, Bromley, Broughton, Burkhardt, Cadwallader, Calloway, Carmichael, Carrington, Cavanaugh, Chadwick, Chamberlain, Chilton, Claffey, Claypool, Clifton, Coffey, Colfax, Colquitt, Conway, Copley, Cotswold, Creighton, Crenshaw, Crowder, Culpepper, Cunningham, Dallimore, Darlington, Davenport, Delaney, Devlin, Doolittle, Dover, Driscoll, Dudley, Dunleavy, Eldridge, Elston, Fairfax, Farnsworth, Fitzgerald, Fitzroy, Flanders, Fleetwood, Gainsborough, Gatling, Goddard, Goodwin, Granger, Greenfield, Griffiths, Harcourt, Hargrove, Harkness, Haverford, Hawkins, Hawthorne, Heathcote, Holbrook, Hollingworth, Holloway, Holmes, Holtz, Howland, Ingles, Jardine, Kenworthy, Kingsley, Langford, Latham, Lathrop, Lockhart, Lodge, Loxley, Lyndon, MacAlister, MacGregor, Mansfield, Marston, Mather, Middleton, Millington, Milton, Montague, Montgomery, Montoya, Morgenthal, Mortimer, Nash, Newcomb, Newkirk, Nightingale, Norwood, Oakley, Ormsby, Osborne, Overton, Pemberton, Pennington, Percival, Pickering, Prescott, Prichard, Quimby, Radcliffe, Rafferty, Rainier, Ramsay, Rawlins, Renshaw, Ridley, Rivers, Rockwell, Roosevelt, Rothschild, Rutherford, Sanderson, Sedgwick, Selwyn, Severance, Sheffield, Sheridan, Sherwood, Shields, Sinclair, Slater, Somerset, Standish, Stanton, Stoddard, Stokes, Stratford, Strickland, Sutherland, Sutton, Talmadge, Tanner, Tennyson, Thackeray, Thatcher, Thorne, Thurston, Tilden, Townsend, Trent, Trevelyan, Trumbull, Underhill, Vanderbilt, Vandermeer, Vickers, Wadsworth, Wakefield, Walpole, Waring, Warwick, Weatherford, Webster, Wharton, Whittaker, Wickham, Wiggins, Wilcox, Winslow, Winthrop, Wolcott, Woodruff, Wycliffe, Yardley, Yates, Yeats, Yule, Zeller, Zimmerman
            \\ \hline
        \end{tabular}
    }
    \caption{All possible values generated for the first name, middle name and last name.}
    \label{tab:names}
\end{table*}
\begin{table*}
    \centering
    \resizebox{\textwidth}{!}{
        \begin{tabular}{lp{21cm}}
        \hline
            \textbf{Relation} & \textbf{Possible Attributes} \\ \hline
            City & "Princeton, NJ", "New York, NY", "Los Angeles, CA", "Chicago, IL", "Houston, TX", "Phoenix, AZ", "Philadelphia, PA", "San Antonio, TX", "San Diego, CA", "Dallas, TX", "San Jose, CA", "Austin, TX", "Jacksonville, FL", "Fort Worth, TX", "Columbus, OH", "San Francisco, CA", "Charlotte, NC", "Indianapolis, IN", "Seattle, WA", "Denver, CO", "Washington, DC", "Boston, MA", "El Paso, TX", "Nashville, TN", "Detroit, MI", "Oklahoma City, OK", "Portland, OR", "Las Vegas, NV", "Memphis, TN", "Louisville, KY", "Baltimore, MD", "Milwaukee, WI", "Albuquerque, NM", "Tucson, AZ", "Fresno, CA", "Mesa, AZ", "Sacramento, CA", "Atlanta, GA", "Kansas City, MO", "Colorado Springs, CO", "Miami, FL", "Raleigh, NC", "Omaha, NE", "Long Beach, CA", "Virginia Beach, VA", "Oakland, CA", "Minneapolis, MN", "Tulsa, OK", "Arlington, TX", "Tampa, FL", "New Orleans, LA", "Wichita, KS", "Cleveland, OH", "Bakersfield, CA", "Aurora, CO", "Anaheim, CA", "Honolulu, HI", "Santa Ana, CA", "Riverside, CA", "Corpus Christi, TX", "Lexington, KY", "Stockton, CA", "Henderson, NV", "Saint Paul, MN", "St. Louis, MO", "Cincinnati, OH", "Pittsburgh, PA", "Greensboro, NC", "Anchorage, AK", "Plano, TX", "Lincoln, NE", "Orlando, FL", "Irvine, CA", "Newark, NJ", "Toledo, OH", "Durham, NC", "Chula Vista, CA", "Fort Wayne, IN", "Jersey City, NJ", "St. Petersburg, FL", "Laredo, TX", "Madison, WI", "Chandler, AZ", "Buffalo, NY", "Lubbock, TX", "Scottsdale, AZ", "Reno, NV", "Glendale, AZ", "Gilbert, AZ", "Winston-Salem, NC", "North Las Vegas, NV", "Norfolk, VA", "Chesapeake, VA", "Garland, TX", "Irving, TX", "Hialeah, FL", "Fremont, CA", "Boise, ID", "Richmond, VA", "Baton Rouge, LA", "Spokane, WA", "Des Moines, IA", "Tacoma, WA", "San Bernardino, CA", "Modesto, CA", "Fontana, CA", "Santa Clarita, CA", "Birmingham, AL", "Oxnard, CA", "Fayetteville, NC", "Moreno Valley, CA", "Rochester, NY", "Glendale, CA", "Huntington Beach, CA", "Salt Lake City, UT", "Grand Rapids, MI", "Amarillo, TX", "Yonkers, NY", "Aurora, IL", "Montgomery, AL", "Akron, OH", "Little Rock, AR", "Huntsville, AL", "Augusta, GA", "Port St. Lucie, FL", "Grand Prairie, TX", "Columbus, GA", "Tallahassee, FL", "Overland Park, KS", "Tempe, AZ", "McKinney, TX", "Mobile, AL", "Cape Coral, FL", "Shreveport, LA", "Frisco, TX", "Knoxville, TN", "Worcester, MA", "Brownsville, TX", "Vancouver, WA", "Fort Lauderdale, FL", "Sioux Falls, SD", "Ontario, CA", "Chattanooga, TN", "Providence, RI", "Newport News, VA", "Rancho Cucamonga, CA", "Santa Rosa, CA", "Peoria, AZ", "Oceanside, CA", "Elk Grove, CA", "Salem, OR", "Pembroke Pines, FL", "Eugene, OR", "Garden Grove, CA", "Cary, NC", "Fort Collins, CO", "Corona, CA", "Springfield, MO", "Jackson, MS", "Alexandria, VA", "Hayward, CA", "Clarksville, TN", "Lancaster, CA", "Lakewood, CO", "Palmdale, CA", "Salinas, CA", "Hollywood, FL", "Pasadena, TX", "Sunnyvale, CA", "Macon, GA", "Pomona, CA", "Escondido, CA", "Killeen, TX", "Naperville, IL", "Joliet, IL", "Bellevue, WA", "Rockford, IL", "Savannah, GA", "Paterson, NJ", "Torrance, CA", "Bridgeport, CT", "McAllen, TX", "Mesquite, TX", "Syracuse, NY", "Midland, TX", "Pasadena, CA", "Murfreesboro, TN", "Miramar, FL", "Dayton, OH", "Fullerton, CA", "Olathe, KS", "Orange, CA", "Thornton, CO", "Roseville, CA", "Denton, TX", "Waco, TX", "Surprise, AZ", "Carrollton, TX", "West Valley City, UT", "Charleston, SC", "Warren, MI", "Hampton, VA", "Gainesville, FL", "Visalia, CA", "Coral Springs, FL", "Columbia, SC", "Cedar Rapids, IA", "Sterling Heights, MI", "New Haven, CT", "Stamford, CT", "Concord, CA", "Kent, WA", "Santa Clara, CA", "Elizabeth, NJ", "Round Rock, TX", "Thousand Oaks, CA", "Lafayette, LA", "Athens, GA", "Topeka, KS", "Simi Valley, CA", "Fargo, ND"
            \\ \hline
        \end{tabular}
    }
    \caption{All possible attributes generated for \textit{city} relation.}
    \label{tab:city}
\end{table*}
\begin{table*}
    \centering
    \resizebox{\textwidth}{!}{
        \begin{tabular}{lp{21cm}}
        \hline
            \textbf{Relation} & \textbf{Possible Attributes} \\ \hline
            Major &  Accounting, Actuarial Science, Advertising, Aerospace Engineering, African American Studies, Agribusiness, Agricultural Engineering, Agriculture, Agronomy, Animal Science, Anthropology, Applied Mathematics, Architecture, Art History, Arts Management, Astronomy, Astrophysics, Athletic Training, Atmospheric Sciences, Biochemistry, Bioengineering, Biological Sciences, Biology, Biomedical Engineering, Biotechnology, Botany, Broadcast Journalism, Business Administration, Business Analytics, Business Economics, Business Information Systems, Chemical Engineering, Chemistry, Civil Engineering, Classics, Cognitive Science, Communication Studies, Communications, Comparative Literature, Computer Engineering, Computer Science, Construction Management, Counseling, Creative Writing, Criminal Justice, Criminology, Culinary Arts, Cybersecurity, Dance, Data Science, Dietetics, Digital Media, Drama, Earth Sciences, Ecology, Economics, Education, Electrical Engineering, Elementary Education, Engineering Physics, Engineering Technology, English, Entrepreneurship, Environmental Engineering, Environmental Science, Environmental Studies, Exercise Science, Fashion Design, Fashion Merchandising, Film Studies, Finance, Fine Arts, Fisheries and Wildlife, Food Science, Forensic Science, Forestry, French, Game Design, Genetics, Geography, Geology, German, Global Studies, Graphic Design, Health Administration, Health Education, Health Informatics, Health Sciences, Healthcare Management, History, Horticulture, Hospitality Management, Human Development, Human Resources Management, Human Services, Industrial Engineering, Information Systems, Information Technology, Interior Design, International Business, International Relations, Journalism, Kinesiology, Labor Studies, Landscape Architecture, Latin American Studies, Law, Legal Studies, Liberal Arts, Linguistics, Management, Management Information Systems, Marine Biology, Marketing, Mass Communications, Materials Science, Mathematics, Mechanical Engineering, Media Studies, Medical Technology, Medicine, Microbiology, Molecular Biology, Music, Music Education, Music Performance, Neuroscience, Nursing, Nutrition, Occupational Therapy, Oceanography, Operations Management, Optometry, Organizational Leadership, Paleontology, Paralegal Studies, Pharmacy, Philosophy, Photography, Physical Education, Physical Therapy, Physics, Physiology, Political Science, Pre-Dental, Pre-Law, Pre-Med, Pre-Pharmacy, Pre-Veterinary, Psychology, Public Administration, Public Health, Public Policy, Public Relations, Quantitative Analysis, Radiologic Technology, Real Estate, Recreation Management, Religious Studies, Renewable Energy, Respiratory Therapy, Risk Management, Robotics, Rural Studies, Sales, Social Work, Sociology, Software Engineering, Spanish, Special Education, Speech Pathology, Sports Management, Statistics, Supply Chain Management, Sustainability, Telecommunications, Theater, Tourism Management, Toxicology, Transportation, Urban Planning, Veterinary Medicine, Victimology, Video Production, Web Development, Wildlife Conservation, Women's Studies, Zoology
            \\ \hline
        \end{tabular}
    }
    \caption{All possible attributes generated for \textit{major} relation.}
    \label{tab:major}
\end{table*}
\begin{table*}
    \centering
    \resizebox{\textwidth}{!}{
        \begin{tabular}{lp{21cm}}
        \hline
            \textbf{Relation} & \textbf{Possible Attributes} \\ \hline
            Company &  Apple, Microsoft, Amazon, Google, Facebook, Berkshire Hathaway, Visa, Johnson \& Johnson, Walmart, Procter \& Gamble, Nvidia, JPMorgan Chase, Home Depot, Mastercard, UnitedHealth Group, Verizon Communications, Pfizer, Chevron, Intel, Cisco Systems, Merck \& Co., Coca-Cola, PepsiCo, Walt Disney, AbbVie, Comcast, Bank of America, ExxonMobil, Thermo Fisher Scientific, McDonald's, Nike, AT\&T, Abbott Laboratories, Wells Fargo, Amgen, Oracle, Costco Wholesale, Salesforce, Medtronic, Bristol-Myers Squibb, Starbucks, IBM, NextEra Energy, Broadcom, Danaher, Qualcomm, General Electric, Honeywell, Citigroup, Lockheed Martin, Union Pacific, Goldman Sachs, Raytheon Technologies, American Express, Boeing, Texas Instruments, Gilead Sciences, S\&P Global, Deere \& Company, Charles Schwab, Colgate-Palmolive, General Motors, Anthem, Philip Morris International, Caterpillar, Target, Intuitive Surgical, Northrop Grumman, Booking Holdings, ConocoPhillips, CVS Health, Altria Group, Eli Lilly and Company, Micron Technology, Fiserv, BlackRock, American Tower, General Dynamics, Lam Research, Zoetis, Applied Materials, Elevance Health, T-Mobile US, Automatic Data Processing, Marsh \& McLennan, Mondelez International, Kroger, Crown Castle, Cigna, Analog Devices, FedEx, CSX, Uber Technologies, Moderna, Truist Financial, Kraft Heinz, HCA Healthcare, Dominion Energy, Cognizant Technology Solutions, Occidental Petroleum, Regeneron Pharmaceuticals, Freeport-McMoRan, eBay, O'Reilly Automotive, Southern Company, Duke Energy, Sherwin-Williams, PayPal, Nucor, Gartner, AutoZone, Cheniere Energy, ServiceNow, Constellation Brands, Discover Financial, U.S. Bancorp, Public Storage, Aflac, Lennar, Johnson Controls, Tyson Foods, Sempra Energy, Southwest Airlines, Las Vegas Sands, McKesson, Baxter International, KLA Corporation, Monster Beverage, Archer Daniels Midland, Eaton, Paccar, Illumina, Intercontinental Exchange, Clorox, Capital One Financial, Estee Lauder, Hess, Becton Dickinson, Parker-Hannifin, Cummins, Ameriprise Financial, Fidelity National Information Services, State Street, Xilinx, Chipotle Mexican Grill, Expeditors International, Roper Technologies, L3Harris Technologies, M\&T Bank, Alcoa, Live Nation Entertainment, Marriott International, Norfolk Southern, DISH Network, Akamai Technologies, Fortinet, Ball Corporation, Corning, Nordstrom, CMS Energy, Nasdaq, BorgWarner, Liberty Media, Sealed Air, PulteGroup, General Mills, Ross Stores, Hewlett Packard Enterprise, Host Hotels \& Resorts, Hilton Worldwide, Snap-on, Zebra Technologies, Leidos, Lincoln National, Weyerhaeuser, CarMax, Rockwell Automation, Allstate, Entergy, NRG Energy, AutoNation, LyondellBasell, Omnicom Group, HollyFrontier, Western Digital, International Flavors \& Fragrances, Eastman Chemical, Xcel Energy, Xylem, Ansys, IPG Photonics, Digital Realty, First Solar, Jacobs Engineering, Cognex, Ingersoll Rand, Fastenal, Allegion, LKQ, AMETEK, WABCO Holdings, Keysight Technologies
            \\ \hline
        \end{tabular}
    }
    \caption{All possible attributes generated for \textit{company} relation.}
    \label{tab:company}
\end{table*}
\begin{table*}
    \centering
    \resizebox{\textwidth}{!}{
        \begin{tabular}{lp{21cm}}
        \hline
            \textbf{Relation} & \textbf{Possible Attributes} \\ \hline
            University & Massachusetts Institute of Technology, Harvard University, Stanford University, California Institute of Technology, University of Chicago, Princeton University, Columbia University, Yale University, University of Pennsylvania, University of California, Berkeley, University of California, Los Angeles, University of Michigan, Ann Arbor, Duke University, Johns Hopkins University, Northwestern University, New York University, University of California, San Diego, University of Southern California, Cornell University, Rice University, University of California, Santa Barbara, University of Washington, University of Texas at Austin, University of Wisconsin-Madison, University of Illinois at Urbana-Champaign, University of North Carolina at Chapel Hill, Washington University in St. Louis, University of Florida, University of Virginia, Carnegie Mellon University, Emory University, Georgetown University, University of California, Irvine, University of Notre Dame, University of Rochester, Boston College, Boston University, Ohio State University, Pennsylvania State University, University of Miami, Purdue University, University of Minnesota, University of Maryland, Michigan State University, University of Colorado Boulder, University of Pittsburgh, University of Arizona, University of Utah, University of California, Davis, University of Massachusetts Amherst, Indiana University Bloomington, University of Connecticut, University of Iowa, University of Missouri, University of Kansas, University of Kentucky, University of Tennessee, University of Alabama, University of Oklahoma, University of Oregon, University of Nebraska-Lincoln, University of South Carolina, University of New Hampshire, University of Vermont, University of Delaware, University of Rhode Island, University of Arkansas, Auburn University, Baylor University, Brigham Young University, Clemson University, Colorado State University, Drexel University, Florida State University, George Washington University, Howard University, Iowa State University, Kansas State University, Louisiana State University, Marquette University, Mississippi State University, North Carolina State University, Northeastern University, Oklahoma State University, Oregon State University, Rutgers University, San Diego State University, Southern Methodist University, Stony Brook University, Syracuse University, Temple University, Texas A\&M University, Texas Tech University, Tulane University, University of Alabama at Birmingham, University of Central Florida, University of Cincinnati, University of Dayton, University of Denver, University of Georgia, University of Houston, University of Idaho, University of Louisville, University of Maryland, Baltimore County, University of Memphis, University of Mississippi, University of Nevada, Las Vegas, University of New Mexico, University of North Texas, University of San Francisco, University of South Florida, University of Texas at Dallas, University of Toledo, University of Tulsa, University of Wyoming, Villanova University, Virginia Tech, Wake Forest University, West Virginia University, Wichita State University, Worcester Polytechnic Institute, Xavier University, Yeshiva University, American University, Arizona State University, Arkansas State University, Ball State University, Boise State University, Bowling Green State University, Bradley University, California Polytechnic State University, California State University, Long Beach, Central Michigan University, Chapman University, City University of New York, Claremont McKenna College, Clark University, College of William \& Mary, DePaul University, Eastern Michigan University, Fairfield University, Florida Atlantic University, Fordham University, Hofstra University, Illinois Institute of Technology, James Madison University, Loyola Marymount University, Loyola University Chicago, Miami University, Middlebury College, New Jersey Institute of Technology, Northern Arizona University, Northern Illinois University, Pepperdine University, Pomona College, Rensselaer Polytechnic Institute, Rhode Island School of Design, Rollins College, Saint Louis University, San Francisco State University, San Jose State University, Santa Clara University, Seattle University, Seton Hall University, Southern Illinois University, Stevens Institute of Technology, SUNY College of Environmental Science and Forestry, SUNY Polytechnic Institute, Texas Christian University, The New School, Towson University, Trinity College, Trinity University, Tufts University, Union College, University at Albany, University at Buffalo, University of Akron, University of Alabama in Huntsville, University of Alaska Anchorage, University of Alaska Fairbanks, University of Baltimore, University of Bridgeport, University of Central Arkansas, University of Charleston, University of Dayton, University of Detroit Mercy, University of Evansville, University of Hartford, University of La Verne, University of Mary Washington, University of Michigan-Dearborn, University of Michigan-Flint, University of Montana, University of Nebraska Omaha, University of Nevada, Reno, University of North Dakota, University of North Florida, University of Northern Colorado, University of Redlands, University of Richmond, University of Saint Joseph, University of San Diego, University of Scranton, University of Sioux Falls, University of South Alabama, University of Southern Mississippi, University of St. Thomas, University of Tampa, University of the Pacific, University of the Sciences, University of Toledo, University of West Georgia, University of Wisconsin-Eau Claire, University of Wisconsin-Green Bay, University of Wisconsin-La Crosse, University of Wisconsin-Milwaukee, University of Wisconsin-Oshkosh, University of Wisconsin-Platteville, University of Wisconsin-River Falls, University of Wisconsin-Stevens Point, University of Wisconsin-Stout, University of Wisconsin-Superior, University of Wisconsin-Whitewater, Ursinus College, Utah State University, Valparaiso University, Vanderbilt University, Vassar College, Villanova University, Virginia Commonwealth University, Wabash College, Wagner College, Wayne State University, Webster University, Weber State University, Wellesley College, Wentworth Institute of Technology, Wesleyan University, Western Carolina University, Western Kentucky University, Western Michigan University, Western Washington University, Westminster College, Whitman College, Whittier College, Willamette University, Williams College, Wittenberg University, Wofford College, Woodbury University, Wright State University, Xavier University, Yale University, York College of Pennsylvania
            \\ \hline
        \end{tabular}
    }
    \caption{All possible attributes generated for \textit{university} relation.}
    \label{tab:university}
\end{table*}

\end{document}